\documentclass[lettersize,journal]{IEEEtran}
\usepackage{amsmath,amsfonts}
\usepackage{algorithmic}
\usepackage{algorithm}
\usepackage{array}
\usepackage[caption=false,font=normalsize,labelfont=sf,textfont=sf]{subfig}
\usepackage{textcomp}
\usepackage{stfloats}
\usepackage{url}
\usepackage{verbatim}
\usepackage{graphicx}
\usepackage{cite}
\usepackage{multirow}
\usepackage{optidef}
\usepackage{makecell}
\usepackage{tikz}
\usepackage{ifthen}
\usepackage{amsmath,amsthm,amsfonts,bm}
\usepackage{optidef}

\def\eqref#1{equation~\ref{#1}}
\def\1{\bm{1}}

\DeclareMathAlphabet{\mathsfit}{\encodingdefault}{\sfdefault}{m}{sl}
\SetMathAlphabet{\mathsfit}{bold}{\encodingdefault}{\sfdefault}{bx}{n}

\newcommand{\R}{\mathbb{R}}

\newcommand{\seq}[3]{
    \ensuremath{\{#1\}_{#2}^{#3}}
}
\newcommand{\XX}[0]{\mathcal{X}}

\theoremstyle{plain}
\newtheorem{theorem}{Theorem}[section]

\theoremstyle{definition}
\newtheorem{definition}{Definition}[section]

\theoremstyle{remark}

\begin{document}

\newcommand{\volint}[3]{
	\ensuremath{\int_{#1} #2 \,\mathrm{d}V(#3)}
}
\newcommand{\surfint}[3]{
	\ensuremath{\int_{\partial #1} #2 \,\mathrm{d}A(#3)}
}
\newcommand{\calL}{\ensuremath{\mathcal{L}}}
\newcommand{\calX}{\ensuremath{\mathcal{X}}}
\newcommand{\calH}{\ensuremath{\mathcal{H}}}
\newcommand{\calD}{\ensuremath{\mathcal{D}}}
\newcommand{\calHX}{\ensuremath{\mathcal{H}(\mathcal{X})}}
\newcommand{\calDX}{\ensuremath{\mathcal{D}(\mathcal{X})}}
\newcommand{\calOX}{\ensuremath{\mathcal{O}(\Omega)}}
\newcommand{\calW}{\ensuremath{\mathcal{W}}}
\newcommand{\setR}{\ensuremath{\mathbb{R}}}
\newcommand{\setC}{\ensuremath{\mathbb{C}}}
\newcommand{\loss}{\ensuremath{\mathrm{loss}}}

\title{On Continuity of Robust and Accurate Classifiers}

\author{
    % Authors and affiliations are removed for double-blind peer review
    % \IEEEauthorblockN{Anonymous Author(s)}
    \IEEEauthorblockA{
        Ramin Barati, Reza Safabakhsh \& Mohammad Rahmati \\
        Department of Computer Engineering\\
        Amirkabir University of Technology\\
        Tehran, Iran\\
        \texttt{\{ramin.barati,safa,rahmati\}@aut.ac.ir}
    }
}
% \author{Ramin Barati,~\IEEEmembership{Staff,~IEEE,}
        % <-this % stops a space

\markboth{Journal of \LaTeX\ Class Files,~Vol.~14, No.~8, August~2021}%
{Shell \MakeLowercase{\textit{et al.}}: A Sample Article Using IEEEtran.cls for IEEE Journals}


\maketitle

\begin{abstract}
The reliability of a learning model is key to the successful deployment of machine learning in various applications. Creating a robust model, particularly one unaffected by adversarial attacks, requires a comprehensive understanding of the adversarial examples phenomenon. However, it is difficult to describe the phenomenon due to the complicated nature of the problems in machine learning. It has been shown that adversarial training can improve the robustness of the hypothesis. However, this improvement usually comes at the cost of decreased performance on natural samples. Hence, it has been suggested that robustness and accuracy of a hypothesis are at odds with each other. In this paper, we put forth the alternative proposal that it is the continuity of a hypothesis that is incompatible with its robustness and accuracy in many of these scenarios. In other words, a continuous function cannot effectively learn the optimal robust hypothesis. We introduce a framework for a rigorous study of harmonic and holomorphic hypothesis in learning theory terms and provide empirical evidence that continuous hypotheses do not perform as well as discontinuous hypotheses in some common machine learning tasks. From a practical point of view, our results suggests that a robust and accurate learning rule would train different continuous hypotheses for different regions of the domain. From a theoretical perspective, our analysis explains the adversarial examples phenomenon in these situations as a conflict between the continuity of a sequence of functions and its uniform convergence to a discontinuous function. Given that many of the contemporary machine learning models are continuous functions, it is important to theoretically study the continuity of robust and accurate classifiers as it is consequential in their construction, analysis and evaluation. It is important in their construction as discontinuities can lead to artifacts in their approximations. It is necessary to analyze these classifiers as they carry topological information about their domain. It is critical in their evaluation because it is revealing of the ways that performance metrics such as accuracy score can fail on assessing these classifiers.
\end{abstract}


\section{Introduction}
Adversarial examples are amongst the most discussed malicious phenomenon in training robust machine learning models. The phenomenon refers to a situation where a trained model is fooled to return an undesirable output on particular inputs that an adversary carefully crafts. The ease of generating these examples and its abundance in most applications of supervised machine learning have posed serious challenges in adoption of machine learning in a safe and secure manner. For example, the phenomenon would allow a malicious actor to alter the predictions of a cloud-based machine learning model\cite{liu2017delving}, or bypass a face-detection model through makeup\cite{lin2021realworld}. Thus, many studies on the possible causes and effects of this phenomenon have been conducted.

While there is no consensus on the reasons behind the emergence of these examples, many facets of the phenomenon have been revealed. For example, Szegedy et al. show that adversarial perturbations are not random and they generalize to other models\cite{szegedy2013intriguing}, or Goodfellow et al. indicate that linear approximations of the model around a test sample is an effective surrogate for the model in the generation of adversarial examples\cite{goodfellow2014explaining}. A known aspect of adversarial examples is that a hypothesis can be trained on its own adversarial examples and improve in robustness. Many related proposals and improvements have been proposed based on this fact, and adversarial training is regarded as the most effective defense method in the community\cite{bai2021recent}. However, it has been shown that adversarial training suffers from a few shortcomings. One of these shortcomings is that it has been observed that adversarial training has a negative effect on the standard performance of the model in some cases\cite{tsipras2018robustness}. This issue is came to be known as the problem of trade-off between accuracy and robustness and has attracted considerable attention in the community. Moreover, adversarially trained hypotheses do not generalize to unseen attacks\cite{kang2019transfer}. These issues have severely hindered the effectiveness of adversarial training\cite{bai2021recent}. As a result, studies have been conducted on mixing different adversarial attacks in hope of training a more robust hypothesis\cite{maini2020adversarial}. Many ideas in the related literature are implicitly or explicitly based on the assumption that the optimal robust hypothesis, if existing, is bounded and Lipschitz continuous\cite{pmlr-v70-cisse17a,10.1007/978-3-030-13453-2_2,9578940}. There is a sentiment that a complete hypothesis space such as artificial neural networks (ANN) should be able to represent the robust hypothesis given enough training samples\cite{goodfellow2014explaining,raghunathan2019adversarial}, and others are hopeful that an inept understanding of deep ANNs would bring about a solution for the phenomenon\cite{buckner2020nature}.

%However, it has been shown that adversarial training suffers from a few shortcomings. As an example, it has been observed that adversarial training has a negative effect on the standard performance of the model\cite{madry2018towards}. There are conflicting views regarding this observation. A popular explanation is that robustness and accuracy are in opposition, and binary classification problems have been synthesized in which standard accuracy and robustness are at odds with each other\cite{tsipras2018robustness}. There is a study that finds negative correlation between the robustness and standard accuracy of adversarially trained models\cite{su2019robustness}. In contrast, there are others that challenge this idea. For example, there is evidence that some of the standard training sets are separable\cite{yang2020closer}, and that robust linear classifiers appear to be possible\cite{tanay2016boundary}. Moreover, adversarially trained hypotheses do not generalize to unseen attacks\cite{kang2019transfer}. This issue has severely hindered the effectiveness of adversarial training\cite{bai2021recent}. As a result, studies have been conducted on mixing different adversarial attacks in hope of training a more robust hypothesis\cite{maini2020adversarial}.

A popular explanation is that robustness and accuracy are in opposition, and binary classification problems have been synthesized in which inconsistencies inherent in the data distribution make an accurate and robust classifier impossible\cite{tsipras2018robustness}. However, there is evidence that some of the standard training sets are separable\cite{yang2020closer}. In contrast, we argue that it is the continuity of finite ANNs that is a hindrance to learning the robust and accurate classifier. In this paper, we are aiming to describe a framework for the study of the discontinuities of a robust and accurate hypothesis. Our framework consists of two parts. First, we need to establish a basis for the study of continuous functions as a learnable hypothesis space. To this end, we consider learning rules that regularize the norm of the gradient of the hypothesis and describe the trained hypothesis through a partial differential equation. In this manner, we are free to analyze the solution without dealing with complexities that are inherent to ANNs. The second part of the framework is concerned with the design of an experiment that would enable us to measure the effect of discontinuities of the robust and accurate classifier on training continuous classifiers. The main insight of the experiment is that we can find a separable set of adversarial and benign training samples for some real-world datasets in which the trade-off between accuracy and robustness vanishes when we relax the condition of continuity of the trained hypothesis. To make sure that the two training sets are separable, we utilize an adversarial examples detector and collect its true positive and negative samples.

In summary, our contributions are:
\begin{itemize}
    \item We present two canonical smoothing strategies for collections of continuous hypotheses that cover some domain and describe the conditions for feasibility of these strategies in special domains. We then describe how their infeasibility contributes to existence of adversarial regions in continuous classifiers found by risk minimizing learning rules.
    \item We introduce weakly-harmonic learning rules that use the Dirichlet energy of the hypothesis as a reguralizer. By doing so, we are able to apply the methods of variational calculus and turn the learning problem into a partial differential equation. We will use this model to study the effects of regularization of the norm of the gradient of the hypothesis. We further introduce the space of weakly-harmonic hypotheses as an abstraction for a solution to a weakly-harmonic learning problem, and provide a convenient representation for these hypotheses. 
    \item By extending the domain of weakly-harmonic hypotheses to complex variables, we present discontinuities in the optimal hypothesis as an explanation for the existence of adversarial regions in the domain of analytic hypothesis. We introduce the space of holomorphic hypotheses and propose a systematic approach for determining the discontinuity of an optimal classifier.
\end{itemize}

We can summarize the significance of our contributions in two parts. First, our results suggests that a robust and accurate hypothesis is a collection of hypotheses that are specialized for different regions of the domain. We further lay a theoretical basis for constructing such hypotheses. Second, by relating the adversarial regions to established concepts in analysis of holomorphic functions, we open a path to a rigorous analysis of the adversarial examples phenomenon that can freely move between the study of the geometrical and the algebraic properties of the hypothesis.

The rest of the paper is organized as follows. In section \ref{sec:background} we will provide an overview of the issue of the trade-off between accuracy and robustness. In section \ref{sec:problem}, we will present the main theoretical results of the paper, and then continue by applying the framework to real-world image classification problems in section \ref{sec:exp}. We will conclude the paper by describing future research directions in section \ref{sec:conclusion}.

We will assume familiarity with the following terms: \textit{Dirichlet energy, Hypothesis space, Probably Approximately Correct (PAC) learnability, Nonuniform learnability, learning rule, loss function, uniform and pointwise convergence, estimation and approximation error, empirical risk minimization (ERM), series representation, Hilbert spaces} and spaces of \textit{square-integrable} $L^2$ and \textit{Sobolev} functions $H^d$.

%We have provided a brief introduction of these terms in Appendix \ref{sec:background} otherwise.

\section{Background}
\label{sec:background}
The problem of training a robust and accurate binary classifier $h^*:\calX\to\R$ could be formulated as the following:
\begin{argmini}
{h\in\mathcal{F}}{\sup_{x\in\calX}\loss(x,h),}
{\label{eq:Example1}}{h^*=}
\end{argmini}
in which $\loss(x,h)$ is some function that measures the fitness of $h$ and $\mathcal{F}$ is a set of candidate hypotheses that is small in some suitable sense. An algorithm that finds some adversarial example $\Tilde{x}\in\calX$ for which $\loss(\Tilde{x},h)$ is large is called an adversarial attack, and augmenting the training data with adversarial examples in training time constitutes adversarial training. A common scenario in adversarial training is to choose $\calX$ to be a union of similar $\ell^p$ balls that are centered on each training point. The issue of the trade-off between robustness and accuracy is the observation that the restriction of $h^*$ to natural examples is consistently outperformed by classifiers that are not adversarially trained. The trade-off between robustness and accuracy is a well-documented phenomenon. Empirical studies, such as those by Su et al. \cite{su2019robustness}, have observed a negative correlation between the robustness and standard accuracy of adversarially trained models across various applications. Wether these correlations are causal or not is up to debate however. Our results suggest this correlation is superfluous and that the underlying cause is the limitations of continuous functions in representing robust and accurate hypotheses. 

Regularization of gradient norm of the hypothesis is an alternative technique to adversarial training that has been shown to be effective in training robust hypotheses\cite{7373334,barrett2022implicit}. An empirical study showed that regularized artificial neural networks exhibit robustness to transferred adversarial examples and that human volunteers find the misclassifications more interpretable\cite{Ross_Doshi-Velez_2018}. In this light, gradient regularization forms an acceptable ground to build our framework upon.

There are also some approaches that try to detect adversarial examples. An empirical study of different methods of detection can be found in \cite{aldahdooh2022adversarial}. It was shown in \cite{grosse2017statistical} that the distribution of adversarial examples differs substantially from the distribution of natural training samples. This difference was shown by performing statistical testing and was reinforced by the observation that a classifier can be trained to detect the adversarial examples produced by an attack. It was shown in \cite{hendrycks2018baseline} that correctly classified examples tend to have greater maximum softmax probabilities compared to erroneously classified and out-of-distribution examples. We will exploit this observation to find two separable training sets where two different continuous hypotheses trained on each set outperform a single continuous hypothesis on the union in terms of accuracy. 

Tsipras et al. argue that the trade-off between robustness and accuracy is an inherent trait of the data distribution and does not stem from insufficient samples\cite{tsipras2018robustness}. To back their claim, they synthesize a binary classification problem in which one of the features, say $x_1$, is a copy of the true label $y$ which is flipped with some probability, and all other features $x_2,\cdots,x_n$ are copies of $y$ that are contaminated with some normally distributed noise. It is proven then that robust accuracy cannot be higher than some upper bound in such a scenario. In contrast, Yang et al. find that the size of the adversarial perturbations in practice is up to seven times smaller than the smallest distance between training samples of different class\cite{yang2020closer}. They continue to prove that if all training samples with different label are separated by $\ell^\infty$ balls, then a nearest-neighbour classifier could be constructed that is locally Lipschitz continuous, robust and accurate in the $\ell^\infty$ balls around the training samples. They further argue that the locally defined classifiers could be smoothed into a global continuous classifier by averaging. In practice however they observe that regularizing local Lipschitz constant in neural networks negatively impact generalization.

The obstacle to robust classification in the example of Tsipras et al. could be attributed to assignment of inconsistent information to a region of the domain of the classifier, which does not appear to apply to adversarial examples in practice. On the other hand, Yang et al. assert that some ANN should be able to represent the robust and accurate classifier, even though current methods fail to find these functions. In contrast, we argue that existence of locally accurate and robust continuous classifiers does not automatically result in the possibility of a global continuous representation of the optimal decision boundary in the coming section.

\section{Method}
\label{sec:problem}

Consider a domain \(\mathcal{X} \subset \mathbb{R}^n\) covered by a set of open sets, where each open set has an associated continuous hypothesis \(h_k\) that is PAC on that set. A natural question arises: can we construct a continuous hypothesis \(h\) that is PAC on \(\mathcal{X}\) from these \(h_k\)? This challenge can be approached by defining \textit{smoothing problems} to reconcile discrepancies between hypotheses over the intersections of their supports.

\begin{definition}[PAC covering]
A PAC covering of a domain $\Omega\subset\setC^n$ is a set $\seq{(h_k,\Omega_k)}{k=1}{\infty}$ in which $h_k:\Omega_k\to\setC$ is continuous and PAC on compact open domains $\Omega_k\subset\setC^n$ for all $k$, and we have that $\Omega$ is a proper subset of $\bigcup_{k=1}^\infty \Omega_k$.
\end{definition}

\begin{definition}[smoothing problem]
    Consider a finite PAC covering $\seq{(h_k,\Omega_k)}{k=1}{K}$ of $\Omega\subset\setC^n$.

    \begin{enumerate}
        \item The first smoothing problem of $\seq{(h_k,\Omega_k)}{k=1}{K}$ is the problem of finding a set $\seq{g_k}{k=1}{K}$ of continuous functions $g_k:\Omega_k\to\setC$ for which
    \begin{equation}
        h_k(z)-h_l(z)=g_k(z)-g_l(z) \quad z\in\Omega_k\cap\Omega_l,
    \end{equation}
    and the smoothed function
    \begin{equation}
        h(z)=h_k(z)-g_k(z)\quad z\in\Omega_k
    \end{equation}
    is PAC on $\Omega$.

    \item the second smoothing problem of $\seq{(h_k,\Omega_k)}{k=1}{K}$ is the problem of finding a set $\seq{g_k}{k=1}{K}$ of continuous nonvanishing functions $g_k:\Omega_k\to\setC$ for which
    \begin{equation}
        \frac{h_k(z)}{h_l(z)}=\frac{g_k(z)}{g_l(z)} \quad z\in\Omega_k\cap\Omega_l,
    \end{equation}
    and the smoothed function
    \begin{equation}
        h(z)=\frac{h_k(z)}{g_k(z)}\quad z\in\Omega_k,
    \end{equation}
    is PAC on $\Omega$.
    \end{enumerate}    
\end{definition}

These smoothing problems are straightforward if each \(h_k\) coincides on the intersections of the open sets. Each smoothing approach suits particular machine learning applications. For instance, the first smoothing problem, which focuses on maintaining the value proximity of the smoothed function to the functions in the original covering, is more appropriate for regression tasks. For binary classification, however, preserving the geometric position of the decision boundary is crucial. Assuming that the zeros of $h_k$ represent the decision boundary and given that $g_k$ is nonvanishing, the decision boundary position for $h$ remains unchanged in a second smoothing problem. However, the second smoothing problem is more challenging to set up, as the zeros of the PAC covering must overlap at intersections for it to be well-defined.

In this paper, we show that constructing a globally continuous, robust, and accurate classifier from locally continuous ones may not be always possible. Our main result provides a necessary and sufficient condition for the feasibility of robust, accurate, continuous binary classifiers on special domains, where the decision boundary can be locally represented by the zeros of holomorphic (complex analytic) functions.

\begin{theorem}
\label{thm:optimality}
Suppose \(\Omega \subset \setC^n\) is a domain of holomorphy and $M\subset\Omega$ is a properly embedded PAC decision boundary. Furthermore, suppose that a covering of holomorphic functions $\seq{f_k,\Omega_k}{k=1}{K}$ of $M$ exists for which $M\cap\Omega_k=\{z\in\Omega_k\,|\,f_k(z)=0\}$ and $f_k(z)$ is not constant on $\Omega_k$. Then, a continuous PAC binary classifier \(h: \Omega\to\R\) exists if and only if there exists some complex-valued function \(f \in \mathcal{O}(\Omega)\) such that $M=\{z\in\Omega\,|\,f(z)=0\}$.
\end{theorem}

Domains of holomorphy are a class of domains that exhibit boundaries with characteristics that make them amenable to analysis of several complex variables and \(\mathcal{O}(\Omega)\) denotes the set of holomorphic functions on \(\Omega\). While it is theoretically possible to prove a similar theorem without invoking concepts related to complex-valued functions, doing so allows us to present the proof within the framework of learning theory, avoiding the direct involvement of abstract mathematical concepts.

As a consequence of Theorem \ref{thm:optimality}, we cannot assume that robust and accurate classifiers can always be represented by continuous functions, such as finite ReLU-activated multi-layer perceptrons, even when the decision boundary is locally representable by polynomial equations. The main barrier to a continuous representation for submanifolds arises from the topological complexity of $\Omega$. Specifically, if $\Omega$ has holes, certain submanifolds may lack a representation as the zero-set of a continuous function. For example, it has been shown that for some domain $\Omega\subset\setC^2$ and submanifold $M\subset\Omega$, no continuous function exists that vanishes only on $M$\cite[page 250]{krantz2001function}. In such cases, no continuous function could represent $M$ as a decision boundary. This limitation is significant for constructing robust, continuous classifiers, as adversarial training may inadvertently exclude the continuous classifier of the original domain, leading to an observed trade-off between accuracy and robustness.

While the existence of such submanifolds is established, we cannot infer from Theorem \ref{thm:optimality} that robust and accurate decision boundaries do not have a continuous representation in general. Instead, we propose an experiment to explore whether decision boundaries without continuous representation exist in real-world scenarios. Finding evidence for such boundaries would suggest that defining topological conditions for such decision boundaries and a formal analysis of the issue is worthwhile. To this end, we introduce the concept of \textit{continuity bias} in machine learning.

\begin{definition}[Continuity Bias]
Consider a discontinuous target \(t \in L^2(\mathcal{X})\). The continuity bias \(\varepsilon \geq 0\) of \(t\) is defined as
\begin{equation}
\varepsilon = \inf_{h \in C(\mathcal{X})} \sup_{x \in \mathcal{X}} \mathrm{loss}(x, h),
\end{equation}
where \(C(\mathcal{X})\) is the set of continuous functions on \(\mathcal{X}\).
\end{definition}

To study the effects of a nonvanishing continuity bias on the output of a learning rule, we examine sequences of continuous functions that converge to a limit function. In the context of a nonuniformly learnable hypothesis space, our objective is to establish two convergence conditions within this sequence: (1) uniform convergence as the training set size grows and (2) uniform convergence as the complexity of the hypothesis space increases.

\begin{definition}[PAC Sequence]
Consider \(\epsilon, \delta \in (0, 1)\), a compact domain \(\Omega \subset \mathbb{C}^n\), a sequence of PAC learnable hypothesis spaces \(\{\mathcal{F}_k : \Omega_k \subset \Omega \to \mathbb{C}\}_{k=1}^\infty\), with \(\mathcal{F}_k \subseteq \mathcal{F}_{k+1}\), sample complexity functions \(m_k : (0, 1)^2 \to \mathbb{N}\), and a sequence of training sets \(\{S_k\}_{k=1}^\infty\), \(|S_k| \geq m_k(\epsilon, \delta)\), sampled from a random variable \(X_k\) supported on \(\Omega_k\). Then \(\{h_k\}_{k=1}^\infty\) is a PAC sequence if \(h_k\in\mathcal{F}_k\) is the output of an ERM learning rule trained on \(S_k\).
\end{definition}

Since continuous functions are closed under uniform convergence, PAC sequences of continuous hypotheses cannot uniformly converge to a discontinuous function. However, a result by Krantz \cite{Krantz2010OnLO} demonstrates that pointwise converging sequences of functions can exist, potentially misleading naive evaluation metrics by exhibiting uniform convergence over certain open subsets of their domain.

\begin{theorem}[Krantz]
\label{thm:krantz}
    Let $\seq{h_k}{1}{\infty}$ be a sequence of holomorphic functions on $\Omega\subseteq\setC^n$. Assume that the sequence converges pointwise to a limit function $h$ on $\Omega$. Then $h$ is holomorphic on a dense open subset of $\Omega$. Also the convergence is uniform on compact subsets of the open set.
\end{theorem}

This trait is not unique to holomorphic functions; analogous theorems to Theorem \ref{thm:krantz} exist for other function spaces. For example, a similar result by Krantz applies to harmonic functions\cite{Krantz2010OnLO}. Here, we continue to use holomorphic functions to maintain focus and avoid introducing more abstract mathematical concepts and theorems.

Our analysis focuses on the mode of convergence of PAC sequences of continuous classifiers corresponding to sequences of hypothesis spaces:
\[
C(\mathcal{X}_1) \subset C(\mathcal{X}_2) \subset \cdots \subset C(\mathcal{X}), \quad \mathcal{X}_k \subset \mathcal{X}_{k+1} \subset \mathcal{X}.
\]
We assume that the random variable \( X_k \) for all PAC sequences is uniformly distributed over its support. This sequence represents an increasing sequence of spaces, where each function in \( C(\mathcal{X}_j) \) is the restriction of some function in \( C(\mathcal{X}_k) \) to \( \mathcal{X}_j \) when \( j < k \).

Our discussion centers on the insight that if the optimal Bayes classifier \( t \in L^2(\mathcal{X}) \) cannot be represented by a continuous function on \( \mathcal{X}_j \), then the corresponding PAC sequence cannot uniformly converge to $t$ on any larger domain \( \mathcal{X}_k \). For instance, if an optimal Bayes classifier \( t \) for grayscale images over \( \mathcal{X}_{\mathrm{gray}} \) cannot be approximated by continuous functions, then any attempt to approximate some extension of \( t \) continuously on \( \mathcal{X}_{\mathrm{rgb}} \) for color images will also fail since:
\[
C(\mathcal{X}_{\mathrm{gray}}) \subset C(\mathcal{X}_{\mathrm{rgb}}).
\]

Our objective is to construct a PAC covering \( \{(h_k, \mathcal{X}_k)\}_{k=1}^{K} \) for which the second smoothing problem is not feasible. If successful, this demonstrates that any PAC sequence of continuous classifiers with a subsequence corresponding to:
\[
\cdots \subset C\left(\bigcup_{k=1}^K \mathcal{X}_k\right) \subset \cdots
\]
converges only pointwise.

The following sections provide the definitions and theorems necessary for proving Theorem \ref{thm:optimality} and a detailed outline of the experiment mentioned. This section is organized into four parts. The first and second parts introduce a partial differential equation that describes the solutions of ERM learning rules regularizing the norm of the hypothesis gradient, and discuss their holomorphic extension into \(\setC^n\). The third part addresses smoothing problems and the proof of Theorem \ref{thm:optimality}. Finally, the fourth part applies the proposed definitions and theorems by developing a computational method to empirically measure the continuity bias.

\subsection{Weakly-Harmonic Hypothesis Spaces}
\label{sec:diri}
Given that we want to study the effects of regularization of the norm of the gradient during training, we have to limit the hypothesis spaces that we consider to spaces that are suitably integrable in both value and derivatives. To this end, we will make use of the Sobolev spaces $H^d(\calX)$ of functions\cite[chapter~7]{gilbarg2001elliptic}. A key concept in these spaces is the concept of weak derivative. Weak derivative is a relaxation of the (strong) derivative of a function through integration by parts\cite[section~7.3]{gilbarg2001elliptic}. For example, the Heaviside step function is the weak derivative of the ReLU function. The weak derivative would equal the strong derivative when $h$ is differentiable.

\begin{definition}[weakly-harmonic problem]
\label{def:weakly-harmonic learning rule}
A weakly-harmonic problem is the problem of finding a hypothesis $h\in H^1(\calX)$ that minimizes the Lagrangian $\calL$:
\begin{equation}
\label{eqn:weakly hamonic problem}
\calL=\frac{1}{2}E[h]+\mathbb{E}_{X}[\loss(x,h)],
\end{equation}
in which $X$ is some random variable supported on $\calX$.
\end{definition}

$E[h]$ in \eqref{eqn:weakly hamonic problem} is the Dirichlet energy of $h$ and is defined as the mean of the squared Euclidean norm of the gradient of $h$. The weakly-harmonic learning problem is the abstraction that we choose to model the learning rules that regularize the norm of the gradient. A more realistic model would replace the Dirichlet energy in the definition with the weighted Dirichlet energy of $h$. However, doing so would introduce complications in the analysis that are out of the scope of this paper.

The form of the optimization problem in \eqref{eqn:weakly hamonic problem} allows for the use of methods of variational calculus\cite[section~11.5]{gilbarg2001elliptic}. As a result, we can solve for $h$ without having to first assume a representation for $h$.

\begin{theorem}
\label{thm:solution}
Suppose that $h\in H^2(\calX)$ is a solution to a weakly-harmonic learning problem. Then $h$ is a weak solution to the following partial differential equation (PDE):
\begin{equation}
\label{eqn:pde}
\Delta h(x)=p(x)\loss_h(x,h),
\end{equation}
in which $\loss_h$ is the variational derivative of $\loss$ with respect to $h$, $p(x)$ is the probability distribution of $X$ and $\Delta$ is the Laplacian operator.
\end{theorem}

The weak solution is a generalization that makes use of the concept of a weak derivative. The PDE of \eqref{eqn:pde} might not have a unique solution when $\loss$ is not convex. Moreover, we only have access to an empirical distribution for $X$. Nevertheless, we propose that the effect of the right hand side of \eqref{eqn:pde} can be replaced by some unknown function independent of $h$. Informally, we can see that $\Delta h(x)=0$ when either $p(x)$ or $\loss_h(x,h)$ vanish. Thus, the optimal solution would be a harmonic function when possible. Otherwise, the optimal solution would not be harmonic and the effects on the optimal solution could be interpreted as some forcing on some harmonic function.

\begin{definition}[weakly-harmonic hypothesis space]
A weakly-harmonic hypothesis space $\calH(\calX)\subset H^2(\calX)$ is the set of hypotheses $h$ that are solutions to the following partial differential equation: 
\begin{align}
\Delta h(x)&=f(x),\\
\frac{\partial h}{\partial\hat{n}}(x)&=g(x) \quad x\in\partial\calX,
\end{align}
in which $\frac{\partial h}{\partial\hat{n}}(x)$ is the derivative of $h$ in the direction of the normal $\hat{n}$ to the boundary $\partial\calX$ and $f,g\in L^2(\calX)$ are some functions.
\end{definition}

Thus, we can divide the learning of the solution to a weakly-harmonic learning problem to two problems; one for $f$ and one for $g$. In other words, if we assume that $u\in H^2(\calX)$ is a strong solution to a weakly-harmonic learning problem, then $u$ has a representation in $\calH(\calX)$ by choosing $f(x)=\Delta u(x)$ and $g(x)=\frac{\partial u}{\partial\hat{n}}(x)$. There are many methods for approximating the solutions to the PDE of a weakly-harmonic hypothesis space. Here, we will be solving the PDEs with the help of the Green function of the Laplacian operator\cite[chapter~2.4]{gilbarg2001elliptic}.

\begin{definition}[eigenvalue problem of $\calH$]
    Suppose that $\calH(\calX)$ is a weakly-harmonic hypothesis space. Then, its corresponding eigenvalue problem is:
    \begin{align}
        \Delta\varphi(x)&=-\lambda^2\varphi(x),\\
        \frac{\partial\varphi}{\partial\hat{n}}(x)&=0\quad x\in\partial\calX.
    \end{align}
\end{definition}

\begin{theorem}
\label{thm:fundamental harmonic}
Suppose that $\calHX$ is a weakly-harmonic hypothesis space, $\partial\calX$ is $C^2$ and that the eigenvalue problem of $\calHX$ is solvable by separation of variables. Then,
\begin{enumerate}
    \item $\calHX$ is nonuniform learnable.
    \item every $h\in\calHX$ has a series representation in $\seq{\varphi_k}{1}{\infty}$ of eigenfunctions of the eigenvalue problem of $\calH$.
	\item $\volint{\calX}{\nabla\varphi_k(x)\cdot\nabla\varphi_l(x)}{x}=\begin{cases}\lambda_k^2\|\varphi_k\|_{L^2}^2& k=l\\0& o.w.\end{cases}.$
\end{enumerate}
\end{theorem}

$C^2$ is a constraint on the smoothness of $\partial\calX$. Informally, if $\partial\calX$ is $C^2$ then a twice differentiable function $\rho$ exists that is zero for every $x\in\partial\calX$ while $\nabla\rho(x)\neq 0$. Theorem \ref{thm:fundamental harmonic} provides us with the minimum requirements needed to make use of $\calHX$ in learning problems\cite[chapter~1]{shalev2014understanding}. Given a representation and the fact that $\calHX$ is nonuniform learnable, any ERM algorithm would suffice to train a hypothesis. Nevertheless, for a concrete hypothesis space, we have to first choose some $\calX$. Even though the boundary of $[0,1]^n$ is not $C^2$, the eigenvalue problem of $\calH([0,1]^n)$ is solvable by separation of variables. The issue of smoothness of the boundary of the hypercube would not pose a problem in practice and we can safely assume that theorem \ref{thm:fundamental harmonic} applies to $\calH([0,1]^n)$.

\begin{theorem}
\label{thm:varphi poly}
    Consider the domain $\omega=[0,\pi]^n$. The solution to the eigenvalue problem of $\calH(\omega)$ is:
    \begin{equation}
        \varphi_\alpha(x)=\cos(\alpha_1 x_1)\cdots\cos(\alpha_n x_n) \quad \alpha\in\mathbb{N}^n,
    \end{equation}
%    and the corresponding diagonal entries of the tuning matrix of $\seq{\varphi_\alpha}{1}{\infty}$ are $\|\alpha\|_2^2$.
\end{theorem}

Since $[0,1]^n$ can be mapped to $\omega$ by simple scaling, $\calH([0,1]^n)$ and $\calH(\omega)$ are the same as far as we are concerned. Theorem \ref{thm:varphi poly} provides us with a representation for $h\in\calH(\omega)$ and shows that $h\in\calH(\omega)$ is analytic when it is constructed using a finite subset of the eigenfunctions. In the next section, we will give a brief introduction to holomorphic functions and domains of holomorphy and move on to show that any such $h$ has a holomorphic extension into a domain of holomorphy as demanded by Theorem \ref{thm:optimality}.

\subsection{Holomorphic Hypotheses}
\label{sec:holo}
Harmonic functions are those functions that minimize the Dirichlet energy and take some value on the boundary of their domain. However, we cannot assume that the training samples are situated on the boundary of the domain in general. This issue can be circumvented by adding extra dimensions to the training samples. In other words, all training points can be positioned on the boundary of some domain in higher dimensions. Of course that does not mean the restriction of such hypothesis to the original domain would be harmonic. Specifically, if we add an imaginary dimension for every real dimension of $x\in\setR^n$, we can assume that $\setR^n$ is the boundary of a product of $n$ copies of half-planes in $\setC^n$.

\begin{definition}[space of holomorphic hypotheses]
    Suppose that $\Omega\subseteq\mathbb{C}^n$ is an open complex domain. The space of holomorphic hypotheses $\calOX$ is the set of functions $h:\Omega\to\mathbb{C}$ that satisfy the homogeneous Cauchy-Riemann equations in each variable $z_j=x_j+iy_j$:
    \begin{equation}
        \frac{\partial h}{\partial \Bar{z}_j}(x,y) =\frac{1}{2}\big(\frac{\partial h}{\partial x_j}(x,y)+i\frac{\partial h}{\partial y_j}(x,y)\big)=0.
    \end{equation}
    Furthermore, $h$ is harmonic and complex differentiable in every $z_j$.
\end{definition}

An important aspect of holomorphic functions is the domain on which these functions are defined on. There is a concept in the analysis of holomorphic functions of several complex variables called a domain of holomorphy\cite[page~6]{krantz2001function} that captures this phenomenon.

\begin{definition}[domain of holomorphy]
    A domain $\Omega\subseteq\mathbb{C}^n$ is a domain of holomorphy if there do not exist nonempty open sets $\Omega_1$ and $\Omega_2$ with $\Omega_2$ connected, $\Omega_1\subseteq\Omega_2\cap\Omega$, such that for every holomorphic function $h$ on $\Omega$ there is a holomorphic function $h_2$ on $\Omega_2$ such that $h=h_2$ on $\Omega_1$.
\end{definition}

In short, holomorphic functions exist on a domain of holomorphy that cannot be holomorphically extended outside $\Omega$.

%Informally, $\Omega$ is a domain of holomorphy if for every $v\in\partial\Omega$ some function $h_v\in\calOX$ exists that blows up on $v$ and does not continue holomorphically beyond $v$. Equivalently, we can find a holomorphic function for which $h_v(z)=0$ and then $h_v(z)^{-1}$ is a holomorphic function that does blow up on $z$.

%First, the $n$-dimensional poly-disk is a domain of holomorphy. Second, if there is a biholomorphic map between a domain of holomorphy $\Omega_1$ and some other domain $\Omega_2$ then $\Omega_2$ is a domain of holomorphy as well. A biholomorphic map is a holomorphic map with a holomorphic inverse. Third, while every $\Omega\subset\setC$ is a domain of holomorphy, it is not true that every domain $\Omega\subset\setC^n,n>1$ is a domain of holomorphy.

%\begin{definition}[poly-disk $D^n(c,r)$]
%    The poly-disk $D^n(c,r)\subset\setC^n$ centered on $c\in\setC^n$ and with multi-radius $r\in\setR^n_+$ is defined as
%    \begin{equation}
%        D^n(c,r)=\{z\in\setC^n\,|\,|z_j-c_j|<r_j \quad j=1,\cdots,n\}.
%    \end{equation}
%\end{definition}

\begin{definition}[tube domains]
    Consider some domain $\calX\subset\setR^n$. The corresponding tube domain $T_\calX\subseteq\setC^n$ of $\calX$ is defined as:
    \begin{equation}
        T_\calX=\{z\in\setC^n\,|\,\Re[z]\in\calX\}.
    \end{equation}
%    The upper-tube domain $U_\omega\subset\setC^n$ of $\omega$ is defined as:
%    \begin{equation}
%        U_\omega=\{z\in\setC^n\,|\,\Re[z]\in\omega,\Im[z]\in\R^n_+\}.
%    \end{equation}
\end{definition}
$T_\calX$ is a domain of holomorphy when $\calX$ is geometrically convex\cite[theorem~3.5.1]{krantz2001function}.

\begin{theorem}
\label{thm:tw}
    Every eigenfunction $\varphi_\alpha\in\calH(\omega),\omega=[0,\pi]^n$ has a holomorphic extension into $T_\omega$,
    \begin{equation}
        \varphi_\alpha(z)=\cos(\alpha_1 z_1)\cdots\cos(\alpha_n z_n)\quad\alpha\in\mathbb{N}^n,
    \end{equation}
    in which
    \begin{equation}
        \cos(x+iy)=\cos(x)\cosh(y)+i\sin(x)\sinh(y).
    \end{equation}
\end{theorem}

%\begin{theorem}
%\label{thm:psi}
%    Every holomorphic function on the upper-tube domain $U_\omega$ of $\omega=[-\pi,\pi]^n$ has a series representation in $\seq{\psi_\alpha}{1}{\infty}$:
%    \begin{equation}
%    \label{eqn:holo features}
%        \psi_\alpha(x)=e^{i\alpha_1 x_1}\cdots e^{i\alpha_n x_n}=e^{i\alpha\cdot x}, \quad \alpha\in\mathbb{N}^n.
%    \end{equation}
%    Moreover, the tuning matrix of $\seq{\psi_\alpha}{1}{\infty}$ is diagonal and its diagonal entries equal $\|\alpha\|_2^2$.
%\end{theorem}

%Taking a closer look at \eqref{eqn:holo features}, we can see that $\psi_\alpha$ is in fact a polynomial in $z_j=e^{ix_j}$. Thus, $U_\omega$ is biholomorphically equivalent with the polydisk $D^n(0,1)$ and is a domain of holomorphy. Due to periodicity of $e^{ix}$, we can only train the positive half of $\omega$.

\begin{definition}[analytic polyhedra]
\label{def:pi}
    Let $\Omega\subseteq\setC^n, n\geq1$ be an open domain. Let $h:\Omega\to\setC^m$ be a holomorphic function on $\Omega$. Consider $\Pi(h)$ defined as
    \begin{equation}
        \Pi(h)=\{z\in\Omega\,|\,|h_j(z)|<1,j=1,\cdots,m\}.
    \end{equation}
    If $\Pi(h)$ is a proper subset of $\Omega$, then $\Pi(h)$ is called an analytic polyhedra.
\end{definition}

The condition for being a proper subset is required so that the boundary of $\Pi(h)$ is only given by $h$ and not $\Omega$. An analytic polyhedra is a domain of holomorphy\cite[section~3.5.1]{krantz2001function}.

\begin{definition}[analytic set]
    Consider an analytic function $h:\setC^n\to\setC^m$. The analytic set $\mathbf{V}(h)$ of $h$ is the set of points $z\in\setC^n$ on which $h(z)$ vanishes:
    \begin{equation}
        \mathbf{V}(h)=\{z\in\setC^n\,|\,h_j(z)=0,j=1,\cdots,m\}.
    \end{equation}
    Moreover, $\mathbf{V}(h)$ has no interior points and cannot be isolated when $n>1$. It is true that $\mathbf{V}(h)\subset\Pi(h)$ by definition.
\end{definition}

\subsection{Solving smoothing problems}
\label{sec:cont bias}
We now turn our attention to the trade-off between robustness and accuracy in classifiers trained on real-world datasets. Our goal is to determine the conditions under which smoothing problems are solvable. Solving the first and second smoothing problems requires meeting both feasibility and optimality conditions. A smoothing problem is feasible if there exists a sequence $\seq{g_k}{k=1}{K}$ that satisfies the required relation with the corresponding PAC covering. It is optimal if the resulting smoothed function is PAC on $\Omega$.

\begin{theorem}
\label{thm:feasibility}
    Consider a finite holomorphic PAC covering $\seq{(h_k,\Omega_k)}{k=1}{K}$ of a domain of holomorphy $\Omega\subset\setC^n$ which is valid for a smoothing problem.
    \begin{enumerate}
        \item The first smoothing problem is always feasible and has holomorphic solutions.
        \item The second smoothing problem is feasible if and only if it is solvable by holomorphic functions.
    \end{enumerate}
\end{theorem}

We now present a sketch of the proof of Theorem \ref{thm:optimality}. Given that $\Omega$ is a domain of holomorphy and we have a holomorphic cover of $M$, it is possible to define a valid second smoothing problem with a continuous solution $g_k(z)=\frac{f_k(z)}{h(z)}$. Theorem \ref{thm:feasibility} then guarantees the existence of $f$. Conversely, given $f$ we can construct $h(z)=|f(z)|t(z)$ as a continuous PAC classifier where  $t:\Omega\to\{+1,-1\}$ represents the optimal Bayes classifier.

In summary, we propose smoothing problems as a computational and theoretical mean for determining the possibility of continuous PAC hypotheses. On domains of holomorphy, the feasibility of these smoothing problems could be decided through their corresponding holomorphic versions. If a smoothing problem is infeasibile, then no sequence of continuous functions would be able to uniformly converge to the optimal hypothesis. Pointwise convergence is still possible though, and learnable hypothesis spaces exist that could fool uniform convergence tests, such as performance on unseen test samples, by converging uniformly on the support of the distribution of the training samples.

\subsection{Measuring the bias}
\label{sec:measuring}
In this section, we use PAC sequences and smoothing problems to investigate the existence of a continuous PAC classifier for a decision boundary \( M \) in a real-world problem. Building on the discussion at the beginning of Section \ref{sec:problem}, we propose a three-step experiment. Failure at any step renders the experiment inconclusive.

Our aim is to identify separable domains \( \seq{\mathcal{X}_k}{k=1}{K} \subset \mathcal{X} \) for which accurate classifiers \( \seq{h_k}{k=1}{K} \subset C(\mathcal{X}) \) can be found, but where any classifier in \( C\left(\cup_{k=1}^K \mathcal{X}_k\right) \) incurs a higher loss than a discontinuous classifier that switches between \( h_k \) based on the domain of a test sample. This setup can be interpreted as finding an optimal solution to a first smoothing problem with data \( \seq{(h_k,\mathcal{X}_k)}{k=1}{K}\cup\{(0, \mathcal{X})\} \). This first smoothing problem is always feasible since \( \seq{(-h_k,\mathcal{X}_k)}{k=1}{K}\cup\{(0, \mathcal{X})\} \) is a solution, but optimality must still be determined.

While second smoothing problems always yield an optimal classifier when feasible, we cannot directly set up a second smoothing problem, as we lack access to the optimal decision boundary. Nevertheless, the feasibility of the second smoothing problem could be inferred by finding an optimal solution for the first smoothing problem. If no optimal solutions exist, then it must be that the second smoothing problem for the optimal decision boundary is infeasible.

First, we need to define suitable domains \( \{\mathcal{X}_k\}_{k=1}^{K} \). Theorem \ref{thm:optimality} does not prescribe how these domains should be chosen. However, if we identify a domain of benign samples \( \mathcal{X}_{\mathrm{ben}} \subset \mathcal{X} \), represented by a set of benign training points \( S_{\mathrm{ben}} \subset \mathcal{X}_{\mathrm{ben}} \), where the target optimal classifier \( t \in L^2(\mathcal{X}_{\mathrm{ben}}) \) can be approximated by a holomorphic function \( h_o \) on the tube domain of \( \mathcal{X}_{\mathrm{ben}} \), we can leverage its analytic polyhedra \( \Pi(h_o) \) as a detector for adversarial examples to construct a suitable covering. Since \( \Pi(h_o) \) is a domain of holomorphy, the optimal holomorphic hypothesis on this domain might not be holomorphically extendable outside \( \Pi(h_o) \).

To ensure the significance of \( \Pi(h_o) \), we require that \( \mathbf{V}(h_o) \) lies on the decision boundary. We will collect two sets of samples: \( S_{\mathrm{easy}} \), comprising benign training samples outside \( \Pi(h_o) \), and \( S_{\mathrm{hard}} \), consisting of adversarial examples that fall inside \( \Pi(h_o) \). These sets delineate two domains \( \mathcal{X}_{\mathrm{easy}} \) and \( \mathcal{X}_{\mathrm{hard}} \), separated by \( \partial\Pi(h_o) \). Alternatively, when the overlap between \( \mathcal{X}_{\mathrm{ben}} \) and \( \mathcal{X}_{\mathrm{hard}} \) is negligible, \( \mathcal{X}_{\mathrm{easy}} \) can be replaced by \( \mathcal{X}_{\mathrm{ben}} \). The proof fails if \( h_o \) proves robust or if either set remains empty.

Next, we quantify the continuity bias \( \varepsilon_t \) of \( t \) over \( \mathcal{X}_{\mathrm{easy}} \cup \mathcal{X}_{\mathrm{hard}} \). We train classifiers \( h, f, g \in C(\mathcal{X}) \) on \( S_{\mathrm{easy}}, S_{\mathrm{hard}}, \) and \( S_{\mathrm{easy}} \cup S_{\mathrm{hard}} \), respectively. A classifier switching between \( h \) and \( f \) based on whether \( x \in \mathcal{X}_{\mathrm{easy}} \) or \( x \in \mathcal{X}_{\mathrm{hard}} \) would be discontinuous on \( \mathcal{X}_{\mathrm{easy}} \cup \mathcal{X}_{\mathrm{hard}} \), whereas \( g \) represents the optimal continuous classifier on this domain. We compare the loss of the discontinuous classifier with \( g \) on \( S_{\mathrm{easy}} \cup S_{\mathrm{hard}} \) as a surrogate for \( \varepsilon_t \) over \( \mathcal{X}_{\mathrm{easy}} \cup \mathcal{X}_{\mathrm{hard}} \). By ensuring \( \mathcal{X}_{\mathrm{easy}} \cap \mathcal{X}_{\mathrm{hard}} = \emptyset \) initially, any bias observed cannot be attributed to inconsistencies between \( S_{\mathrm{easy}} \) and \( S_{\mathrm{hard}} \).

Finally, we select an arbitrary hypothesis \( q \in C(\mathcal{X}) \) and measure \( \varepsilon_q \) for a regression problem over \( \mathcal{X}_{\mathrm{easy}} \cup \mathcal{X}_{\mathrm{hard}} \). This step is necessary because we are not solving the smoothing problem directly in the previous step. If \( \varepsilon_q > 0 \), it suggests that the observation regarding \( \varepsilon_t \) might not be solely due to the discontinuity of \( t \), thereby causing the proof to fail. Conversely, a \( \varepsilon_q \) of zero completes the proof, confirming \( \varepsilon_t > 0 \).

However, at present, practical representation and training of all continuous or holomorphic hypotheses on a domain is not possible, necessitating reliance on empirical measurements based on finite representations. Additionally, when dealing with more than two classes, we extend our analysis to all binary classification problems resulting from combinations of these classes. If our hypothesis holds true, we should observe \( \varepsilon_t > 0 \) for some or all of these binary classification problems.

To establish \( \varepsilon > 0 \), we will employ a method of statistical hypothesis testing \cite[section~3.5]{lehmann2005testing}. This involves defining a null hypothesis \( \mathbf{H_0} \) and an alternative hypothesis \( \mathbf{H_1} \). The goal of statistical hypothesis testing is to determine if our observations provide sufficient evidence to reject \( \mathbf{H_0} \) in favor of \( \mathbf{H_1} \). In our context, the null hypothesis is \( \mathbf{H_0}: \varepsilon \leq 0 \), and the alternative hypothesis is \( \mathbf{H_1}: \varepsilon > 0 \).

Since we are testing the average of the continuity bias measurements over a batch of samples, we assume that the distribution of \( \varepsilon \) is normal with some unknown mean and variance. To conduct the test, we first compute the sample mean \( \bar{\varepsilon} \) and the unbiased sample standard deviation \( \sigma \), and then calculate the test statistic
\[
T = \frac{\bar{\varepsilon}}{\sigma} \sqrt{n},
\]
where \( n \) is the size of the samples of \( \varepsilon \). We then compare \( T \) with the critical value
\[
t = t_{\nu}^{-1}(1 - c),
\]
where \( t_{\nu}^{-1} \) is the inverse cumulative distribution function of a t-student random variable with \( \nu = n - 1 \) degrees of freedom, and \( c \in (0, 1) \) is the confidence level of the test. We will accept \( \mathbf{H_1} \) if \( T > t \), and reject it otherwise. Notably, rejecting \( \mathbf{H_1} \) does not imply acceptance of \( \mathbf{H_0} \).

\section{Experiments}
\label{sec:exp}
We have conducted three experiments in order to verify and test the proposal as described in section \ref{sec:measuring}. All three experiments were conducted on four datasets of the MNIST family; the MNIST dataset itself\cite{lecun-mnisthandwrittendigit-2010}, FashionMNIST\cite{xiao2017fashionmnist}, Kuzushiji-MNIST\cite{clanuwat2018deep}, and EMNIST\cite{DBLP:journals/corr/CohenATS17}. The MNIST family contains $28\times28$ gray-scale images of categories such as handwritten digits everyday clothing items. We have reported the results for Kuzushiji-MNIST (KMNIST) and EMNIST in the appendix accompanied by extra details regarding the experiments. To demonstrate a case of failure, we will discuss how the first step of the measurement process fails on CIFAR10\cite{alex2009learning}.

The loss function was the (binary) cross-entropy loss and the optimizer used in the experiments was an implementation of batch gradient decent that is provided by the PyTorch(2.1.0)\cite{NEURIPS2019_9015} library. Each batch contained 32 training samples and we continued the training for 5 epochs. The learning rate of the optimizer was set to 0.001, the momentum was 0.9 and the weight decay was 1. All experiments were conducted on an Intel Core i5-6600K @ 3.5GHz, 8.00 GB of RAM and no GPUs. The implementation of the PGD attack was provided by the CleverHans(4.0.0)\cite{papernot2018cleverhans} library and the TorchAttacks(3.5.1)\cite{kim2020torchattacks} library was used for a CPU friendly implementation of AutoAttack\cite{croce2020reliable}. In both cases the attacks were constrained in a $\ell^\infty$-ball with a radius of 0.3. The PGD attack performed 40 steps of size of 0.01 and the Square attack performed 5000 queries. We have left the value of the other parameters of the attacks to their default in their respective implementations. The detection scores are computed with the help of the Scikit-learn(1.3.0)\cite{scikit-learn} library.

We have made use of a series representation and a convolutional architecture to deal with the complexity that arises from the combinatorial explosion of eigenfunctions of the Laplace operator in high dimensions. Furthermore, to position the analytic set of holomorphic classifiers on the the decision boundary as required in the first part of the experiment, we will make use of the Fourier basis,
\begin{equation}
\label{eqn:holo features}
        \psi_\alpha(x)=e^{i\alpha_1 x_1}\cdots e^{i\alpha_n x_n}=e^{i\alpha\cdot x}, \quad \alpha\in\mathbb{N}^n,
\end{equation}
in the aforementioned representation, and further proceed to take the complex logarithm,
\begin{equation}
\log(z)=\ln(|z|)+i\angle z,
\end{equation}
of the hypothesis before passing its real part to the softmax function,
\begin{equation}
    \mathbb{P}(\mathrm{class}=j|z)=\frac{|h_j(z)|}{\sum_{k=1}^m|h_k(z)|},\quad j=1,\cdots,m,
\end{equation}
the details of which are available in the appendix. It could be inspected that $\psi_\alpha$ is a solution to the eigenvalue problem of the Laplace operator on $[-\pi,\pi]^n$ but with periodic boundary conditions. Nevertheless, the part of the experiment that is concerned with actual computation of continuity bias is performed using simple multi-layer perceptron (MLP) neural networks.

\subsection{Adversarial example detection}
In this experiment, we test whether we can detect an adversarial example using the analytic polyhedra $\Pi(h_o)$ of the best holomorphic hypothesis that we can find over $\calX_{\mathrm{ben}}$. We have performed two versions of this experiment. In one of the experiments the attacker is aware that we want to capture the adversarial examples in $\Pi(h_o)$ and in the other experiment the attacker is not aware of this issue. We have repeated each experiment 10 times and report the average of the scores in Table \ref{tbl:attack summary}. We can see in the table that the precision score of $\Pi(h_o)$ is high, which is consistent with the assumption that $\calX_\mathrm{hard}$ has little overlap with the support of natural samples. The recall score indicates that $\Pi(h_o)$ covers a considerable portion of the adversarial region of $h_o$. Thus, we argue that the evidence is in favor of assuming that the adversarial examples that fall inside $\Pi(h_o)$ cannot assign inconsistent data to a subset of the domain.

\begin{table*}[]
\setlength\extrarowheight{2.5pt}
\caption{The performance of the analytic polyhedra in detecting adversarial examples. The results are divided by the property whether the attacker is aware of the extra class or not.}
\label{tbl:attack summary}
\centering
\begin{tabular}{c|cccccc|cccccc|}
\cline{2-13}
                                     & \multicolumn{6}{c|}{FMNIST}                                                                                                                                      & \multicolumn{6}{c|}{MNIST}                                                                                                                                       \\ \hline
\multicolumn{1}{|c|}{\multirow{2}{*}{attack target}}         & \multicolumn{3}{c|}{w/ extra}                                                             & \multicolumn{3}{c|}{w/o extra}                                       & \multicolumn{3}{c|}{w/ extra}                                                             & \multicolumn{3}{c|}{w/o extra}                                       \\ \cline{2-13}
\multicolumn{1}{|c|}{}         & \multicolumn{1}{c|}{precision} & \multicolumn{1}{c|}{recall} & \multicolumn{1}{c|}{F1}    & \multicolumn{1}{c|}{precision} & \multicolumn{1}{c|}{recall} & F1    & \multicolumn{1}{c|}{precision} & \multicolumn{1}{c|}{recall} & \multicolumn{1}{c|}{F1}    & \multicolumn{1}{c|}{precision} & \multicolumn{1}{c|}{recall} & F1    \\ \hline
\multicolumn{1}{|c|}{untargeted}     & \multicolumn{1}{c|}{0.934}     & \multicolumn{1}{c|}{0.458}  & \multicolumn{1}{c|}{0.614} & \multicolumn{1}{c|}{0.936}     & \multicolumn{1}{c|}{0.471}  & 0.625 & \multicolumn{1}{c|}{0.908}     & \multicolumn{1}{c|}{0.409}  & \multicolumn{1}{c|}{0.562} & \multicolumn{1}{c|}{0.910}     & \multicolumn{1}{c|}{0.422}  & 0.575 \\ 
\hline
\end{tabular}
\end{table*}

\subsection{Infeasibility of holomorphic classifiers}
As explained in section \ref{sec:cont bias}, we trained three holomorphic classifiers $h,f,g$ on $S_{\mathrm{ben}}$, $S_{\mathrm{hard}}$ and $S_{\mathrm{ben}}\cup S_{\mathrm{hard}}$ respectively. We then compared the accuracy of the three classifiers on test sets consisting of unseen samples from $S_{\mathrm{ben}}$ and $S_{\mathrm{hard}}$. The results are reported in Table \ref{tbl:essential}. We can see that the evidence points to the fact that the robust and accurate holomorphic classifier could be constructed by patching together $h$ and $f$, but not with a single holomorphic hypothesis $g$. The drop in relative accuracy of $g$ with respect to $h$ and $f$ on MNIST are $1\%$ and $3\%$, respectively, which agrees with the $2\%$ drop as reported by Tsipiras et. al.\cite{tsipras2018robustness}. Furthermore, we can see that $S_{\mathrm{ben}}$ is adversarial with respect to $f$ which goes against intuition since $S_{\mathrm{hard}}$ is closer to the true decision boundary of $S_{\mathrm{ben}}$ and should act as its support vectors. For completeness, we note that our attempts to train a holomorphic PAC classifier on CIFAR10 were unsuccessful. Furthermore, the trained classifier proved robust against the PGD attack, as we were unable to generate any adversarial examples. Therefore, we refrain from making any assertions regarding the continuity bias on CIFAR10.

To provide a better description of the performance of the hypotheses, we attacked $h_o$ using unseen attacks from AutoAttack suit of adversarial attacks, untargeted and targeted APGD(-T)\cite{croce2021mind}, FAB\cite{10.5555/3524938.3525143} and Square\cite{DBLP:conf/eccv/AndriushchenkoC20}, and computed the performance of adversarial detection of $\Pi(h_o)$ and classification accuracy of $f$ on the detected adversarial examples. We have reported the results in Table \ref{tbl:autoattack}. We can see that the performance of $f$ on detected adversarial examples is similar, with the exception of the Square attack. The detection rate of $\Pi(h_o)$ is significantly better against the Square attack on the other hand. The decrease of performance against targeted attacks is expected as the attack would tend to evade $\Pi(h_o)$. The difference of the results on the Square attack is probably due to its non-reliance on gradient information and the specific manner in which adversarial examples are constructed in this attack. The increase of detection accuracy in conjunction with the decrease in classification accuracy for the case of Square attack however might indicate that $\Pi(h_o)$ is not topologically simple.

We emphasize that the results of the experiment is not indicative of the robust performance of the hypothesis against attacks that have access to $h$, $f$ and the switching mechanism simultaneously. If the attacker has access to both hypotheses, both could be evaded. This observation is not in conflict with our conclusion that continuous functions are not adequate, but suggests that there are more than one cluster of continuously inseparable regions in the neighborhood of natural samples.

\begin{table}[]
\setlength\extrarowheight{2.5pt}
\caption{The accuracy of holomorphic classifiers on adversarial $S_\mathrm{hard}$ and benign $S_\mathrm{ben}$ test samples. $h$ is the holomorphic hypothesis that is trained on $S_\mathrm{ben}$, $f$ is the one that is trained on $S_\mathrm{hard}$ and $g$ is the classifier that is trained on both training sets.}
\label{tbl:essential}
\centering
\begin{tabular}{c|cc|cc|cc|}
\cline{2-7}
                             & \multicolumn{2}{c|}{FMNIST}               & \multicolumn{2}{c|}{MNIST} & \multicolumn{2}{c|}{CIFAR10}                \\ \hline
\multicolumn{1}{|c|}{target} & \multicolumn{1}{c|}{hard} & benign & \multicolumn{1}{c|}{hard} & benign & \multicolumn{1}{c|}{hard} & benign \\ \hline
\multicolumn{1}{|c|}{$h$}    & \multicolumn{1}{c|}{0.000}       & 0.755  & \multicolumn{1}{c|}{0.000}       & 0.890 & \multicolumn{1}{c|}{-}       & 0.250 \\
\multicolumn{1}{|c|}{$f$}    & \multicolumn{1}{c|}{0.752}       & 0.361  & \multicolumn{1}{c|}{0.900}       & 0.302 & \multicolumn{1}{c|}{-}       & - \\
\multicolumn{1}{|c|}{$g$}    & \multicolumn{1}{c|}{0.741}       & 0.728  & \multicolumn{1}{c|}{0.880}       & 0.886 & \multicolumn{1}{c|}{-}       & - \\ \hline
\end{tabular}
\end{table}

% \usepackage{multirow}
\begin{table}[]
\setlength\extrarowheight{2.5pt}
\caption{The performance of $\Pi(h_o)$ and $f$ on unseen attacks.}
\label{tbl:autoattack}
\begin{tabular}{c|c|c|c|c|c|}
\cline{2-6}
                                                                                                             & dataset & APGD  & APGD-T & FAB & Square \\ \hline
\multicolumn{1}{|c|}{\multirow{2}{*}{\begin{tabular}[c]{@{}c@{}}$\Pi(h_o)$ detection\\ accuracy\end{tabular}}} & MNIST   & 0.251 & 0.361  & 0.414 & 0.622  \\ \cline{2-2}
\multicolumn{1}{|c|}{}                                                                                       & FMNIST  & 0.302 & 0.282  & 0.286 & 0.600  \\ \hline
\multicolumn{1}{|c|}{\multirow{2}{*}{\begin{tabular}[c]{@{}c@{}}$f$ classification\\ accuracy\end{tabular}}} & MNIST   & 0.862 & 0.790  & 0.819 & 0.507  \\ \cline{2-2}
\multicolumn{1}{|c|}{}                                                                                       & FMNIST  & 0.736 & 0.689  & 0.738 & 0.414  \\ \hline
\end{tabular}
\end{table}

\subsection{Measuring the continuity bias}
In this experiment we will be using $784\times 512\times 512\times 256\times 1$ ReLU activated MLPs to measure $\varepsilon$. Since we do not need to regularize the norm of the gradient in this step, we did not apply any weight decay in training ANNs. We have used random $784\times 512\times 512\times 1$ ReLU activated MLPs as arbitrary continuous targets for the third step of the experiment. We repeated the tests 20 times and computed the mean and standard deviation of the sampled continuity biases. Finally, we calculated the required test statistic and compared it to the corresponding critical value with the confidence level of 0.01. The results of the tests is reported in Table \ref{tbl:h0}. The results in case of discontinuous targets shows that we can accept $\mathbf{H_1}$ with a very high confidence. $\mathbf{H_1}$ is rejected in case of continuous targets as predicted.

% \usepackage{multirow}
\begin{table}[]
\setlength\extrarowheight{2.5pt}
\caption{Results of the statistical hypothesis testing of $\varepsilon>0$. The results show that $\varepsilon\leq0$ cannot be rejected in case of the regression tasks. In classification tasks on the other hand, we can accept $\varepsilon>0$ with high confidence.}
\label{tbl:h0}
\centering
\begin{tabular}{|c|c|cc|cc|}
\hline
\multirow{2}{*}{dataset} & \multirow{2}{*}{\thead{critical \\ value}} & \multicolumn{2}{c|}{\thead{continuous \\ target}}                             & \multicolumn{2}{c|}{\thead{discontinuous \\ target}}                     \\ \cline{3-6} 
                         &                                 & \multicolumn{1}{c|}{\thead{test \\ statistic}} & $\mathbf{H_1}$                & \multicolumn{1}{c|}{\thead{test \\ statistic}} & $\mathbf{H_1}$            \\ \hline
MNIST                    & 2.54                            & \multicolumn{1}{c|}{2.10}          & rejected & \multicolumn{1}{c|}{10.77}           & accepted \\ \hline
FMNIST                   & 2.54                            & \multicolumn{1}{c|}{1.41}           & rejected & \multicolumn{1}{c|}{5.21}           & accepted \\ \hline
\end{tabular}
\end{table}

\section{Conclusion}
\label{sec:conclusion}
In this paper, we examined the impact of target discontinuities in machine learning problems on the output of learning algorithms. To the best of our knowledge, we are the first to provide empirical evidence suggesting that it may be impossible to train a hypothesis that is simultaneously robust, accurate, and continuous in real-world scenarios. A limitation of our experimental approach is the difficulty in isolating the effects of specific choices made in constructing the hypotheses. As a result, we acknowledge that the question of continuity for the optimal robust hypothesis remains open until an appropriate proof is established. We have drawn parallels and compared our approach with recent findings in the appendix.

From the perspective of weakly-harmonic learning, our results suggest that the strong solution of the corresponding PDE may not exist. This implies that robust learning rules over continuous functions must address domain adaptation. Test-Time Training (TTT) \cite{pmlr-v119-sun20b} and Deep Mixture-of-Experts (DMoE) \cite{eigen2014learning} are promising approaches. While TTT can adapt weak PDE formulations to robust learning, DMoE methods can leverage domains of holomorphy. Approximating weak PDE solutions remains an open problem \cite{zang2020weak, doi:10.1137/22M1488405}, which we leave for future work. Independently, several authors have found evidence that MoE models can bypass the robustness-accuracy trade-off\cite{10374121,10552117,zhang2025optimizingrobustnessaccuracymixture}. 

There are also improvements that could be applied to the methodology of the paper. As stated in section \ref{sec:exp}, we could not perform the experiment on color images. Nevertheless, we believe that improvements in this regard also hang on a suitable method of domain decomposition which would allow for finer holomorphic PAC coverings to be incorporated in the measurement of continuity bias on more sophisticated classification problems.

\bibliographystyle{IEEEtran}
\bibliography{references}

\begin{thebibliography}{10}
\providecommand{\url}[1]{#1}
\csname url@samestyle\endcsname
\providecommand{\newblock}{\relax}
\providecommand{\bibinfo}[2]{#2}
\providecommand{\BIBentrySTDinterwordspacing}{\spaceskip=0pt\relax}
\providecommand{\BIBentryALTinterwordstretchfactor}{4}
\providecommand{\BIBentryALTinterwordspacing}{\spaceskip=\fontdimen2\font plus
\BIBentryALTinterwordstretchfactor\fontdimen3\font minus
  \fontdimen4\font\relax}
\providecommand{\BIBforeignlanguage}[2]{{%
\expandafter\ifx\csname l@#1\endcsname\relax
\typeout{** WARNING: IEEEtran.bst: No hyphenation pattern has been}%
\typeout{** loaded for the language `#1'. Using the pattern for}%
\typeout{** the default language instead.}%
\else
\language=\csname l@#1\endcsname
\fi
#2}}
\providecommand{\BIBdecl}{\relax}
\BIBdecl

\bibitem{liu2017delving}
\BIBentryALTinterwordspacing
Y.~Liu, X.~Chen, C.~Liu, and D.~Song, ``Delving into transferable adversarial
  examples and black-box attacks,'' in \emph{International Conference on
  Learning Representations}, 2017. [Online]. Available:
  \url{https://openreview.net/forum?id=Sys6GJqxl}
\BIBentrySTDinterwordspacing

\bibitem{lin2021realworld}
C.-S. Lin, C.-Y. Hsu, P.-Y. Chen, and C.-M. Yu, ``Real-world adversarial
  examples via makeup,'' in \emph{ICASSP 2022 - 2022 IEEE International
  Conference on Acoustics, Speech and Signal Processing (ICASSP)}, 2022, pp.
  2854--2858.

\bibitem{szegedy2013intriguing}
C.~Szegedy, W.~Zaremba, I.~Sutskever, J.~Bruna, D.~Erhan, I.~J. Goodfellow, and
  R.~Fergus, ``Intriguing properties of neural networks,'' in \emph{2nd
  International Conference on Learning Representations, {ICLR} 2014, Banff, AB,
  Canada, April 14-16, 2014, Conference Track Proceedings}, Y.~Bengio and
  Y.~LeCun, Eds., 2014.

\bibitem{goodfellow2014explaining}
I.~J. Goodfellow, J.~Shlens, and C.~Szegedy, ``Explaining and harnessing
  adversarial examples,'' in \emph{3rd International Conference on Learning
  Representations, {ICLR} 2015, San Diego, CA, USA, May 7-9, 2015, Conference
  Track Proceedings}, Y.~Bengio and Y.~LeCun, Eds., 2015.

\bibitem{bai2021recent}
\BIBentryALTinterwordspacing
T.~Bai, J.~Luo, J.~Zhao, B.~Wen, and Q.~Wang, ``Recent advances in adversarial
  training for adversarial robustness,'' in \emph{Proceedings of the Thirtieth
  International Joint Conference on Artificial Intelligence, {IJCAI-21}}, Z.-H.
  Zhou, Ed.\hskip 1em plus 0.5em minus 0.4em\relax International Joint
  Conferences on Artificial Intelligence Organization, 8 2021, pp. 4312--4321,
  survey Track. [Online]. Available:
  \url{https://doi.org/10.24963/ijcai.2021/591}
\BIBentrySTDinterwordspacing

\bibitem{tsipras2018robustness}
D.~Tsipras, S.~Santurkar, L.~Engstrom, A.~Turner, and A.~Madry, ``Robustness
  may be at odds with accuracy,'' in \emph{International Conference on Learning
  Representations}, 2019.

\bibitem{kang2019transfer}
D.~Kang, Y.~Sun, T.~Brown, D.~Hendrycks, and J.~Steinhardt, ``Transfer of
  adversarial robustness between perturbation types,'' 2019.

\bibitem{maini2020adversarial}
P.~Maini, E.~Wong, and J.~Z. Kolter, ``Adversarial robustness against the union
  of multiple perturbation models,'' in \emph{Proceedings of the 37th
  International Conference on Machine Learning}, ser. ICML'20.\hskip 1em plus
  0.5em minus 0.4em\relax JMLR.org, 2020.

\bibitem{pmlr-v70-cisse17a}
M.~Cisse, P.~Bojanowski, E.~Grave, Y.~Dauphin, and N.~Usunier, ``Parseval
  networks: Improving robustness to adversarial examples,'' in
  \emph{Proceedings of the 34th International Conference on Machine Learning},
  ser. Proceedings of Machine Learning Research, D.~Precup and Y.~W. Teh, Eds.,
  vol.~70.\hskip 1em plus 0.5em minus 0.4em\relax PMLR, 06--11 Aug 2017, pp.
  854--863.

\bibitem{10.1007/978-3-030-13453-2_2}
T.~Huster, C.-Y.~J. Chiang, and R.~Chadha, ``Limitations of the lipschitz
  constant as a defense against adversarial examples,'' in \emph{ECML PKDD 2018
  Workshops}, C.~Alzate, A.~Monreale, H.~Assem, A.~Bifet, T.~S. Buda,
  B.~Caglayan, B.~Drury, E.~Garc{\'i}a-Mart{\'i}n, R.~Gavald{\`a},
  I.~Koprinska, S.~Kramer, N.~Lavesson, M.~Madden, I.~Molloy, M.-I. Nicolae,
  and M.~Sinn, Eds.\hskip 1em plus 0.5em minus 0.4em\relax Cham: Springer
  International Publishing, 2019, pp. 16--29.

\bibitem{9578940}
A.~Fan, X.~Jiang, Y.~Ma, X.~Mei, and J.~Ma, ``Smoothness-driven consensus based
  on compact representation for robust feature matching,'' \emph{IEEE
  Transactions on Neural Networks and Learning Systems}, vol.~34, no.~8, pp.
  4460--4472, 2023.

\bibitem{raghunathan2019adversarial}
\BIBentryALTinterwordspacing
A.~Raghunathan*, S.~M. Xie*, F.~Yang, J.~Duchi, and P.~Liang, ``Adversarial
  training can hurt generalization,'' in \emph{ICML 2019 Workshop on
  Identifying and Understanding Deep Learning Phenomena}, 2019. [Online].
  Available: \url{https://openreview.net/forum?id=SyxM3J256E}
\BIBentrySTDinterwordspacing

\bibitem{buckner2020nature}
C.~Buckner, ``Understanding adversarial examples requires a theory of artefacts
  for deep learning,'' \emph{Nature Machine Intelligence}, vol.~2, pp.
  731--736, 12 2020.

\bibitem{yang2020closer}
\BIBentryALTinterwordspacing
Y.-Y. Yang, C.~Rashtchian, H.~Zhang, R.~R. Salakhutdinov, and K.~Chaudhuri, ``A
  closer look at accuracy vs. robustness,'' in \emph{Advances in Neural
  Information Processing Systems}, H.~Larochelle, M.~Ranzato, R.~Hadsell,
  M.~Balcan, and H.~Lin, Eds., vol.~33.\hskip 1em plus 0.5em minus 0.4em\relax
  Curran Associates, Inc., 2020, pp. 8588--8601. [Online]. Available:
  \url{https://proceedings.neurips.cc/paper_files/paper/2020/file/61d77652c97ef636343742fc3dcf3ba9-Paper.pdf}
\BIBentrySTDinterwordspacing

\bibitem{su2019robustness}
D.~Su, H.~Zhang, H.~Chen, J.~Yi, P.-Y. Chen, and Y.~Gao, ``Is robustness the
  cost of accuracy? -- a comprehensive study on the robustness of 18 deep image
  classification models,'' in \emph{Computer Vision -- ECCV 2018}, V.~Ferrari,
  M.~Hebert, C.~Sminchisescu, and Y.~Weiss, Eds.\hskip 1em plus 0.5em minus
  0.4em\relax Cham: Springer International Publishing, 2018, pp. 644--661.

\bibitem{7373334}
C.~Lyu, K.~Huang, and H.-N. Liang, ``A unified gradient regularization family
  for adversarial examples,'' in \emph{2015 IEEE International Conference on
  Data Mining}, 2015, pp. 301--309.

\bibitem{barrett2022implicit}
\BIBentryALTinterwordspacing
D.~Barrett and B.~Dherin, ``Implicit gradient regularization,'' in
  \emph{International Conference on Learning Representations}, 2021. [Online].
  Available: \url{https://openreview.net/forum?id=3q5IqUrkcF}
\BIBentrySTDinterwordspacing

\bibitem{Ross_Doshi-Velez_2018}
A.~Ross and F.~Doshi-Velez, ``Improving the adversarial robustness and
  interpretability of deep neural networks by regularizing their input
  gradients,'' \emph{Proceedings of the AAAI Conference on Artificial
  Intelligence}, vol.~32, no.~1, Apr. 2018.

\bibitem{aldahdooh2022adversarial}
A.~Aldahdooh, W.~Hamidouche, S.~A. Fezza, and O.~D{\'e}forges, ``Adversarial
  example detection for dnn models: A review and experimental comparison,''
  \emph{Artificial Intelligence Review}, vol.~55, no.~6, pp. 4403--4462, 2022.

\bibitem{grosse2017statistical}
K.~{Grosse}, P.~{Manoharan}, N.~{Papernot}, M.~{Backes}, and P.~{McDaniel},
  ``{On the (Statistical) Detection of Adversarial Examples},'' \emph{arXiv
  e-prints}, p. arXiv:1702.06280, Feb. 2017.

\bibitem{hendrycks2018baseline}
\BIBentryALTinterwordspacing
D.~Hendrycks and K.~Gimpel, ``A baseline for detecting misclassified and
  out-of-distribution examples in neural networks,'' in \emph{International
  Conference on Learning Representations}, 2017. [Online]. Available:
  \url{https://openreview.net/forum?id=Hkg4TI9xl}
\BIBentrySTDinterwordspacing

\bibitem{krantz2001function}
S.~Krantz, \emph{Function Theory of Several Complex Variables}, ser. AMS
  Chelsea Publishing Series.\hskip 1em plus 0.5em minus 0.4em\relax American
  Mathematical Society, 2001.

\bibitem{Krantz2010OnLO}
S.~G. Krantz, ``On limits of sequences of holomorphic functions,'' \emph{Rocky
  Mountain Journal of Mathematics}, vol.~43, pp. 273--283, 2010.

\bibitem{gilbarg2001elliptic}
D.~Gilbarg and N.~Trudinger, \emph{Elliptic Partial Differential Equations of
  Second Order}, ser. Classics in Mathematics.\hskip 1em plus 0.5em minus
  0.4em\relax Springer Berlin Heidelberg, 2001.

\bibitem{shalev2014understanding}
S.~Shalev-Shwartz and S.~Ben-David, \emph{Understanding Machine Learning: From
  Theory to Algorithms}, ser. Understanding Machine Learning: From Theory to
  Algorithms.\hskip 1em plus 0.5em minus 0.4em\relax Cambridge University
  Press, 2014.

\bibitem{lehmann2005testing}
E.~L. Lehmann and J.~P. Romano, \emph{Testing statistical hypotheses}, 3rd~ed.,
  ser. Springer Texts in Statistics.\hskip 1em plus 0.5em minus 0.4em\relax New
  York: Springer, 2005.

\bibitem{lecun-mnisthandwrittendigit-2010}
Y.~LeCun and C.~Cortes, ``{MNIST} handwritten digit database,'' 2010.

\bibitem{xiao2017fashionmnist}
H.~Xiao, K.~Rasul, and R.~Vollgraf, ``Fashion-mnist: a novel image dataset for
  benchmarking machine learning algorithms,'' 2017, cite
  arxiv:1708.07747Comment: Dataset is freely available at
  https://github.com/zalandoresearch/fashion-mnist Benchmark is available at
  http://fashion-mnist.s3-website.eu-central-1.amazonaws.com/.

\bibitem{clanuwat2018deep}
T.~Clanuwat, M.~Bober-Irizar, A.~Kitamoto, A.~Lamb, K.~Yamamoto, and D.~Ha.
  (2018) Deep learning for classical japanese literature.

\bibitem{DBLP:journals/corr/CohenATS17}
\BIBentryALTinterwordspacing
G.~Cohen, S.~Afshar, J.~Tapson, and A.~van Schaik, ``{EMNIST:} an extension of
  {MNIST} to handwritten letters,'' \emph{CoRR}, vol. abs/1702.05373, 2017.
  [Online]. Available: \url{http://arxiv.org/abs/1702.05373}
\BIBentrySTDinterwordspacing

\bibitem{alex2009learning}
K.~Alex, ``Learning multiple layers of features from tiny images,''
  \emph{https://www. cs. toronto. edu/kriz/learning-features-2009-TR. pdf},
  2009.

\bibitem{NEURIPS2019_9015}
A.~Paszke, S.~Gross, F.~Massa, A.~Lerer, J.~Bradbury, G.~Chanan, T.~Killeen,
  Z.~Lin, N.~Gimelshein, L.~Antiga, A.~Desmaison, A.~Kopf, E.~Yang, Z.~DeVito,
  M.~Raison, A.~Tejani, S.~Chilamkurthy, B.~Steiner, L.~Fang, J.~Bai, and
  S.~Chintala, ``Pytorch: An imperative style, high-performance deep learning
  library,'' in \emph{Advances in Neural Information Processing Systems
  32}.\hskip 1em plus 0.5em minus 0.4em\relax Curran Associates, Inc., 2019,
  pp. 8024--8035.

\bibitem{papernot2018cleverhans}
N.~Papernot, F.~Faghri, N.~Carlini, I.~Goodfellow, R.~Feinman, A.~Kurakin,
  C.~Xie, Y.~Sharma, T.~Brown, A.~Roy, A.~Matyasko, V.~Behzadan,
  K.~Hambardzumyan, Z.~Zhang, Y.-L. Juang, Z.~Li, R.~Sheatsley, A.~Garg,
  J.~Uesato, W.~Gierke, Y.~Dong, D.~Berthelot, P.~Hendricks, J.~Rauber, and
  R.~Long, ``Technical report on the cleverhans v2.1.0 adversarial examples
  library,'' \emph{arXiv preprint arXiv:1610.00768}, 2018.

\bibitem{kim2020torchattacks}
H.~Kim, ``Torchattacks: A pytorch repository for adversarial attacks,''
  \emph{arXiv preprint arXiv:2010.01950}, 2020.

\bibitem{croce2020reliable}
F.~Croce and M.~Hein, ``Reliable evaluation of adversarial robustness with an
  ensemble of diverse parameter-free attacks,'' in \emph{ICML}, 2020.

\bibitem{scikit-learn}
F.~Pedregosa, G.~Varoquaux, A.~Gramfort, V.~Michel, B.~Thirion, O.~Grisel,
  M.~Blondel, P.~Prettenhofer, R.~Weiss, V.~Dubourg, J.~Vanderplas, A.~Passos,
  D.~Cournapeau, M.~Brucher, M.~Perrot, and E.~Duchesnay, ``Scikit-learn:
  Machine learning in {P}ython,'' \emph{Journal of Machine Learning Research},
  vol.~12, pp. 2825--2830, 2011.

\bibitem{croce2021mind}
F.~Croce and M.~Hein, ``Mind the box: $l_1$-apgd for sparse adversarial attacks
  on image classifiers,'' in \emph{ICML}, 2021.

\bibitem{10.5555/3524938.3525143}
------, ``Minimally distorted adversarial examples with a fast adaptive
  boundary attack,'' in \emph{Proceedings of the 37th International Conference
  on Machine Learning}, ser. ICML'20.\hskip 1em plus 0.5em minus 0.4em\relax
  JMLR.org, 2020.

\bibitem{DBLP:conf/eccv/AndriushchenkoC20}
\BIBentryALTinterwordspacing
M.~Andriushchenko, F.~Croce, N.~Flammarion, and M.~Hein, ``Square attack: {A}
  query-efficient black-box adversarial attack via random search,'' in
  \emph{Computer Vision - {ECCV} 2020 - 16th European Conference, Glasgow, UK,
  August 23-28, 2020, Proceedings, Part {XXIII}}, ser. Lecture Notes in
  Computer Science, A.~Vedaldi, H.~Bischof, T.~Brox, and J.~Frahm, Eds., vol.
  12368.\hskip 1em plus 0.5em minus 0.4em\relax Springer, 2020, pp. 484--501.
  [Online]. Available: \url{https://doi.org/10.1007/978-3-030-58592-1\_29}
\BIBentrySTDinterwordspacing

\bibitem{pmlr-v119-sun20b}
\BIBentryALTinterwordspacing
Y.~Sun, X.~Wang, Z.~Liu, J.~Miller, A.~Efros, and M.~Hardt, ``Test-time
  training with self-supervision for generalization under distribution
  shifts,'' in \emph{Proceedings of the 37th International Conference on
  Machine Learning}, ser. Proceedings of Machine Learning Research, H.~D. III
  and A.~Singh, Eds., vol. 119.\hskip 1em plus 0.5em minus 0.4em\relax PMLR,
  13--18 Jul 2020, pp. 9229--9248. [Online]. Available:
  \url{https://proceedings.mlr.press/v119/sun20b.html}
\BIBentrySTDinterwordspacing

\bibitem{eigen2014learning}
D.~Eigen, M.~Ranzato, and I.~Sutskever, ``Learning factored representations in
  a deep mixture of experts,'' 2014.

\bibitem{zang2020weak}
Y.~Zang, G.~Bao, X.~Ye, and H.~Zhou, ``Weak adversarial networks for
  high-dimensional partial differential equations,'' \emph{Journal of
  Computational Physics}, vol. 411, p. 109409, 2020.

\bibitem{doi:10.1137/22M1488405}
\BIBentryALTinterwordspacing
F.~Chen, J.~Huang, C.~Wang, and H.~Yang, ``Friedrichs learning: Weak solutions
  of partial differential equations via deep learning,'' \emph{SIAM Journal on
  Scientific Computing}, vol.~45, no.~3, pp. A1271--A1299, 2023. [Online].
  Available: \url{https://doi.org/10.1137/22M1488405}
\BIBentrySTDinterwordspacing

\bibitem{10374121}
H.~Aboutalebi, M.~J. Shafiee, C.-E.~A. Tai, and A.~Wong, ``Knowing is half the
  battle: Enhancing clean data accuracy of adversarial robust deep neural
  networks via dual-model bounded divergence gating,'' \emph{IEEE Access},
  vol.~12, pp. 48\,174--48\,188, 2024.

\bibitem{10552117}
X.~Wei, S.~Zhao, and B.~Li, ``Revisiting the trade-off between accuracy and
  robustness via weight distribution of filters,'' \emph{IEEE Transactions on
  Pattern Analysis and Machine Intelligence}, vol.~46, no.~12, pp. 8870--8882,
  2024.

\bibitem{zhang2025optimizingrobustnessaccuracymixture}
\BIBentryALTinterwordspacing
X.~Zhang, K.~Xu, Z.~Hu, and R.~Wang, ``Optimizing robustness and accuracy in
  mixture of experts: A dual-model approach,'' 2025. [Online]. Available:
  \url{https://arxiv.org/abs/2502.06832}
\BIBentrySTDinterwordspacing

\bibitem{10197639}
E.~Dobriban, H.~Hassani, D.~Hong, and A.~Robey, ``Provable tradeoffs in
  adversarially robust classification,'' \emph{IEEE Transactions on Information
  Theory}, vol.~69, no.~12, pp. 7793--7822, 2023.

\bibitem{10186217}
H.~Taheri, R.~Pedarsani, and C.~Thrampoulidis, ``Asymptotic behavior of
  adversarial training in binary linear classification,'' \emph{IEEE
  Transactions on Neural Networks and Learning Systems}, pp. 1--9, 2023.

\bibitem{10049380}
S.~Kanai, M.~Yamada, H.~Takahashi, Y.~Yamanaka, and Y.~Ida, ``Relationship
  between nonsmoothness in adversarial training, constraints of attacks, and
  flatness in the input space,'' \emph{IEEE Transactions on Neural Networks and
  Learning Systems}, pp. 1--15, 2023.

\bibitem{pmlr-v97-zhang19p}
\BIBentryALTinterwordspacing
H.~Zhang, Y.~Yu, J.~Jiao, E.~Xing, L.~E. Ghaoui, and M.~Jordan, ``Theoretically
  principled trade-off between robustness and accuracy,'' in \emph{Proceedings
  of the 36th International Conference on Machine Learning}, ser. Proceedings
  of Machine Learning Research, K.~Chaudhuri and R.~Salakhutdinov, Eds.,
  vol.~97.\hskip 1em plus 0.5em minus 0.4em\relax PMLR, 09--15 Jun 2019, pp.
  7472--7482. [Online]. Available:
  \url{https://proceedings.mlr.press/v97/zhang19p.html}
\BIBentrySTDinterwordspacing

\bibitem{9586061}
C.~Liu, M.~Salzmann, and S.~Süsstrunk, ``Training provably robust models by
  polyhedral envelope regularization,'' \emph{IEEE Transactions on Neural
  Networks and Learning Systems}, vol.~34, no.~6, pp. 3146--3160, 2023.

\bibitem{ilyas2019adversarial}
A.~Ilyas, S.~Santurkar, D.~Tsipras, L.~Engstrom, B.~Tran, and A.~Madry,
  ``Adversarial examples are not bugs, they are features,'' in \emph{Advances
  in Neural Information Processing Systems 32: Annual Conference on Neural
  Information Processing Systems 2019, NeurIPS 2019, December 8-14, 2019,
  Vancouver, BC, Canada}, H.~M. Wallach, H.~Larochelle, A.~Beygelzimer,
  F.~d'Alch{\'{e}}{-}Buc, E.~B. Fox, and R.~Garnett, Eds., 2019, pp. 125--136.

\bibitem{kumano2024theoretical}
\BIBentryALTinterwordspacing
S.~Kumano, H.~Kera, and T.~Yamasaki, ``Theoretical understanding of learning
  from adversarial perturbations,'' in \emph{The Twelfth International
  Conference on Learning Representations}, 2024. [Online]. Available:
  \url{https://openreview.net/forum?id=Ww9rWUAcdo}
\BIBentrySTDinterwordspacing

\bibitem{bishop2006pattern}
C.~Bishop, \emph{Pattern Recognition and Machine Learning}.\hskip 1em plus
  0.5em minus 0.4em\relax Springer, January 2006.

\end{thebibliography}

\clearpage
\appendices

\section{Parallels and Comparisons}
\label{sec:lit}
We agree with Yang et al.\cite{yang2020closer} that the observed trade-off between accuracy and robustness is not caused by inconsistencies similar to the example that Tsipras et al. provide. In contrast to Yang et al. however, we will not go all the way as to claim that an ANN exists which represents the true decision boundary. Yang et al. are assuming that the existence of a continuous PAC covering is enough to construct a globally continuous PAC hypothesis by smoothing. They reach this conclusion by assuming a first smoothing problem of the PAC covering. While this conclusion may be sound for a regression problem, approaching the issue from the perspective of a second smoothing problem reveals that the situation for classification problems are decisively different.

The proposal of Tsipras et al.\cite{tsipras2018robustness} assumes that the trade-off between accuracy and robustness in classification problems exists because robust and accurate hypotheses are not realizable. In other words, they argue that the learning problem is constrained in such a way that does not allow for robust and accurate hypotheses. In contrast, our proposal does not impose any restrictions on the optimal Bayes classifier and we argue that the observation of the trade-off is a part of the approximation error of the learning algorithm that would vanish if we were to give up on the continuity of the hypotheses.

Dobriban et al.\cite{10197639} consider a similar phenomenon in the case where the margin of a linear binary classifier is given its own class and find out that certain discontinuities in the output of a robust learning rule would appear when the parameters of the learning problem is changing. While both discussions has similarities, they are considering the very different case of discontinuities in the hyper-parameter space of a learning rule.

Taheri et al.\cite{10186217} analyses the behavior of binary classifiers in high dimensions. They find exact asymptotics for standard and adversarial error in the separable case for generalized linear models. Their analysis does not cover topological properties of the data and is focused on the effects of changing different hyper-parameters of adversarial training. Nevertheless, their proposal also studies sequences of optimal classifiers and has similarities with our proposal in that regard.

Kanai et al.\cite{10049380} study the smoothness of the loss of adversarial training. They reach the conclusion that adversarial training is more difficult than standard training because smoothness of the gradient of the loss function with respect to model parameters and the flatness of the loss landscape in the input space exhibit a trade-off. Consequently, they show that smoother adversarial loss functions such as TRADES\cite{pmlr-v97-zhang19p} would show better convergence properties. We emphasize that smoothness of the hypothesis is not the same as the smoothness of the loss landscape. Specifically, the loss landscape of the optimal Bayes classifier can be completely flat in the input space even though the hypothesis itself is not continuous. Thus, their analysis does not take into account the possibility of discontinuity of the optimal hypothesis.

Liu et al.\cite{9586061} propose a certification method that uses polyhedral envelops for better approximation of the robust regions in the neighborhood of samples. While very similar, we should emphasize that the polyhedral envelops considered by Liu et al. are different from the analytic polyhedra that we consider in our experiments. Nevertheless, both methods demonstrate the computational feasibility of approximating nontrivial domains using polyhedra.

Zhang et al.\cite{pmlr-v97-zhang19p} assert that smoothness is an indispensable property of robust models, and introduce the TRADES loss function as a means for the control of the trade-off between accuracy and robustness. This line of reasoning is in direct contrast with our approach. We believe that smoothness is only a convenient \textit{local} property that simplifies approximating PAC hypotheses when it is possible for them to be so. However, smoothness conditions are a burden when we consider learning problems that involve domains that are not topologically simple such as natural images.

Ilyas et al. posit that vulnerable hypotheses are attending nonrobust features in the natural data that are useful in classifying benign samples, but are a hindrance to the robustness of the hypotheses\cite{ilyas2019adversarial}. We believe on the other hand that a better description is that useful features are only defined locally, and that hypotheses attend to these features because no alternatives exist that would perform as good as these features on benign samples.

Kumano et al.\cite{kumano2024theoretical} shed some light on the mechanism of an observation made in Ilyas et al.\cite{ilyas2019adversarial} paper. The observation is that mislabeled adversarial examples can be used to train a hypothesis that is accurate on natural samples. Kumano et al. conclude that  adversarial perturbations can be represented as weighted combinations of training data, and models trained on adversarial examples with incorrect labels learn a decision boundary that is similar to the boundary for the original natural data distribution. In our framework, this observation is analogous to the assertion that adversarial examples are mislabeled because these wrong labels are consistent with the labels of natural samples from the point of view of continuity. Our experimental results, presented in Table \ref{tbl:essential}, shows that if these examples were used for training with the correct label, the natural samples would be mislabeled instead.

Goodfellow et al. suggest that vulnerable artificial neural networks are converging to the optimal linear hypothesis\cite{goodfellow2014explaining}. In contrast, we picture the robust and accurate hypothesis as a collection of analytic functions that cannot be glued together in a way that the hypothesis remain PAC in its domain. In this light, we argue that vulnerable ANNs are converging to the best piecewise linear hypothesis that is defined on the support of the distribution of natural samples.

\section{Details of Evaluations}
\label{sec:details}
In this section we will report the results of performing the experiments on KMNIST and EMNIST, and will provide more details about the experiments in general. KMNIST is a drop-in replacement for the MNIST dataset ($28 \times 28$ grayscale, 70,000 images) of Hiragana characters. EMNIST extends MNIST with English alphabets, and provides more samples for the digits collection. We will be using the digits subset of EMNIST to keep the experiment consistent with other datasets.

Table \ref{tbl:ke-h0} reports the results of measuring the continuity bias for KMNIST and EMNIST. The results agree with what we have observed on MNIST and FMNIST.

\begin{table}[]
\setlength\extrarowheight{2.5pt}
\caption{Results of the statistical hypothesis testing of $\varepsilon>0$. The results show that $\varepsilon\leq0$ cannot be rejected in case of the regression tasks. In classification tasks on the other hand, we can accept $\varepsilon>0$ with high confidence.}
\label{tbl:ke-h0}
\centering
\begin{tabular}{|c|c|cc|cc|}
\hline
\multirow{2}{*}{dataset} & \multirow{2}{*}{\thead{critical \\ value}} & \multicolumn{2}{c|}{\thead{continuous \\ target}}                             & \multicolumn{2}{c|}{\thead{discontinuous \\ target}}                     \\ \cline{3-6} 
                         &                                 & \multicolumn{1}{c|}{\thead{test \\ statistic}} & $\mathbf{H_1}$                & \multicolumn{1}{c|}{\thead{test \\ statistic}} & $\mathbf{H_1}$            \\ \hline
KMNIST                    & 2.54                            & \multicolumn{1}{c|}{1.69}          & rejected & \multicolumn{1}{c|}{5.60}           & accepted \\ \hline
EMNIST                   & 2.54                            & \multicolumn{1}{c|}{2.43}           & rejected & \multicolumn{1}{c|}{7.61}           & accepted \\ \hline
\end{tabular}
\end{table}

\begin{figure}
\centering
\subfloat{\includegraphics[width=0.45\columnwidth]{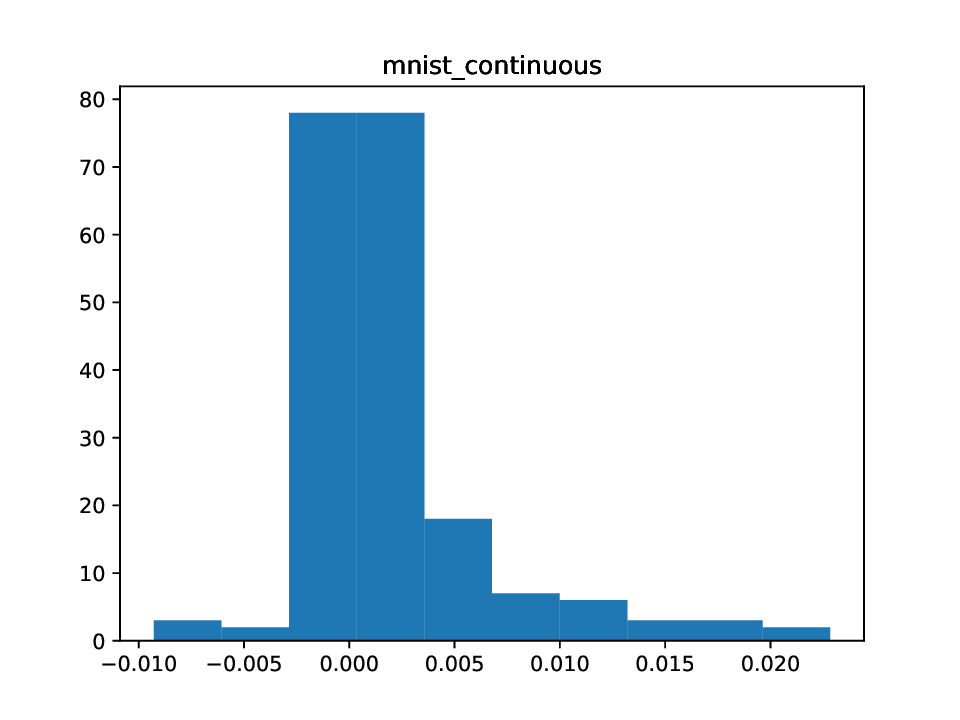}}
\subfloat{\includegraphics[width=0.45\columnwidth]{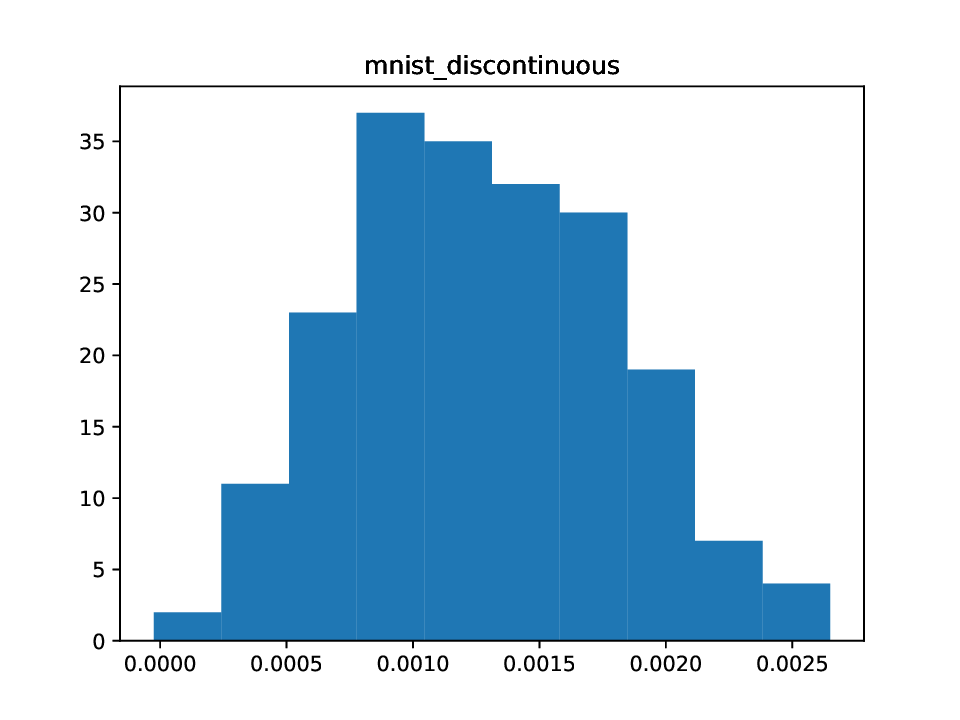}}
\caption{The histogram of the measured continuity bias for MNIST.}
\label{fig:hist_mnist}
\end{figure}

\begin{figure}
\centering
\subfloat{\includegraphics[width=0.45\columnwidth]{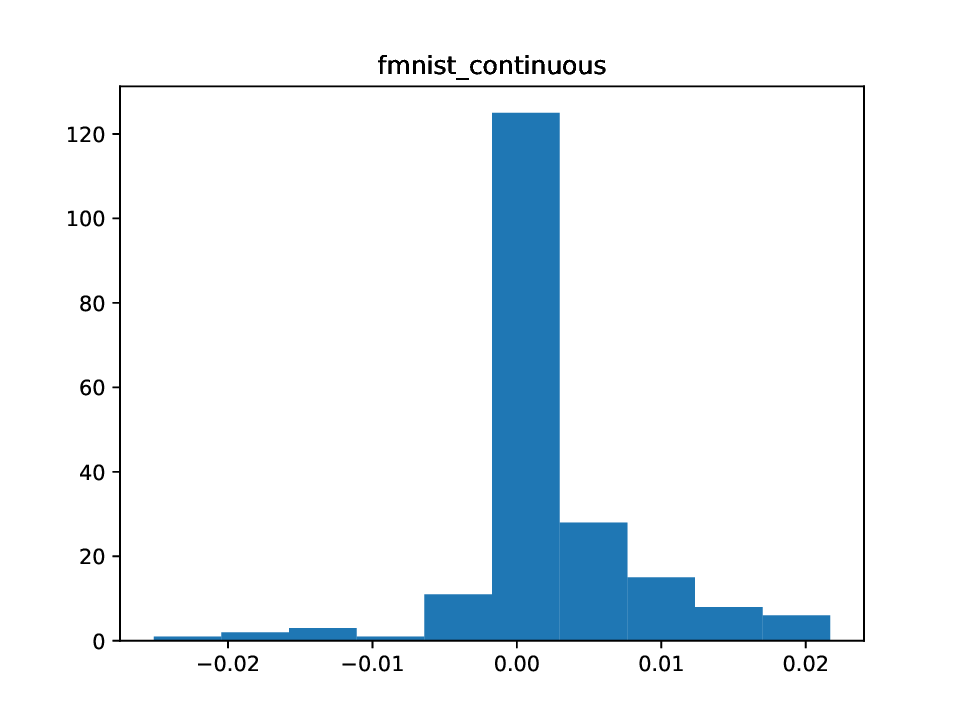}}
\subfloat{\includegraphics[width=0.45\columnwidth]{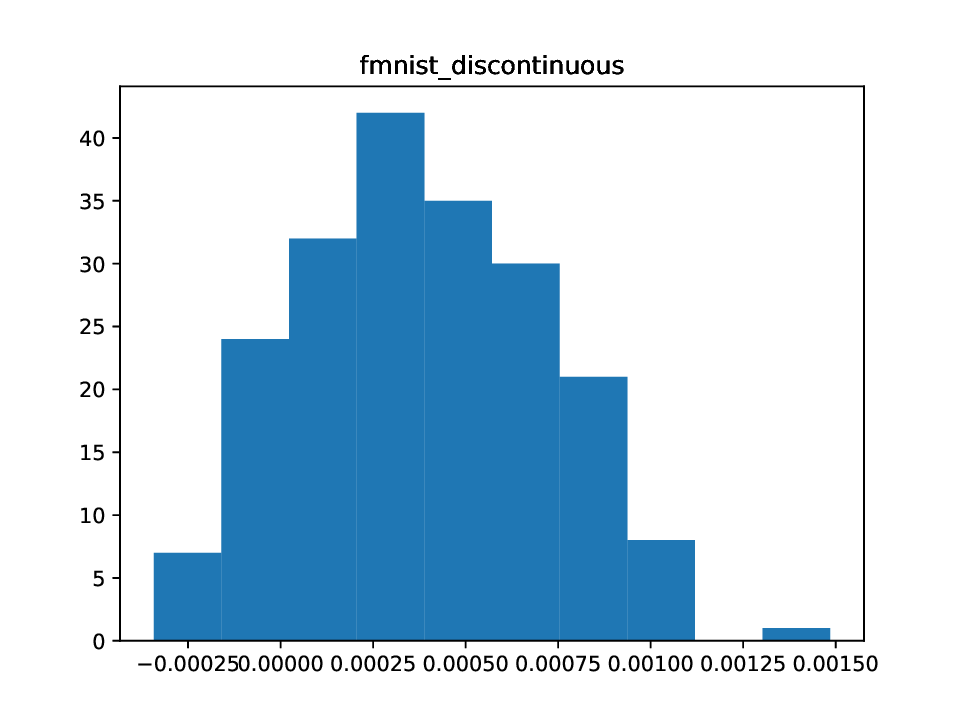}}
\caption{The distribution of the measured continuity bias for FMNIST.}
\label{fig:hist_fmnist}
\end{figure}

\begin{figure}
\centering
\subfloat{\includegraphics[width=0.45\columnwidth]{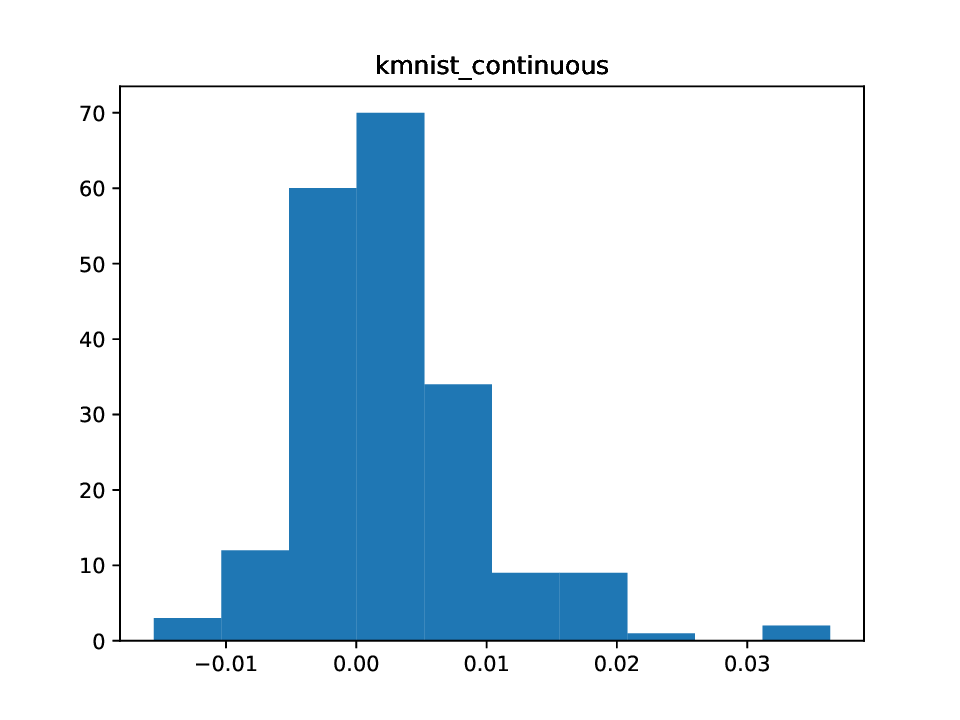}}
\subfloat{\includegraphics[width=0.45\columnwidth]{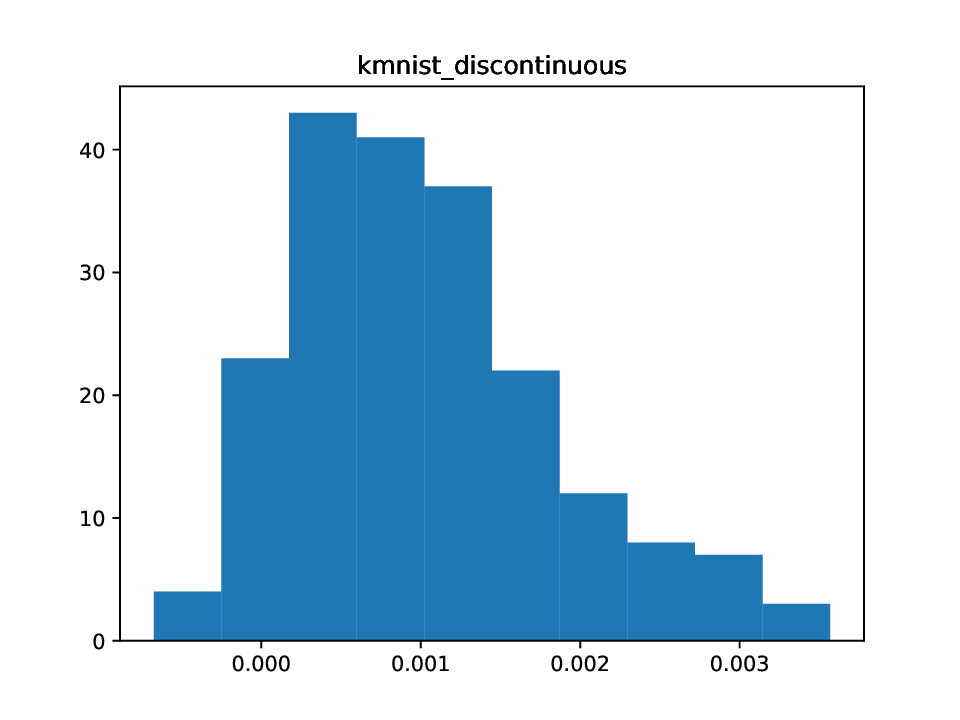}}
\caption{The distribution of the measured continuity bias for KMNIST.}
\label{fig:hist_kmnist}
\end{figure}

\begin{figure}
\centering
\subfloat{\includegraphics[width=0.45\columnwidth]{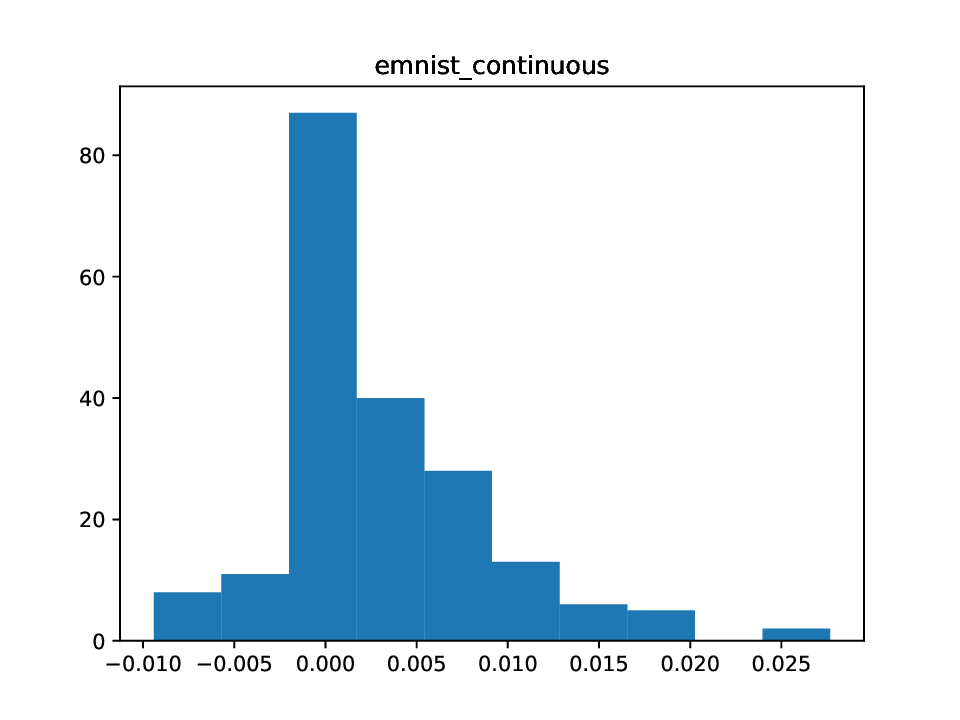}}
\subfloat{\includegraphics[width=0.45\columnwidth]{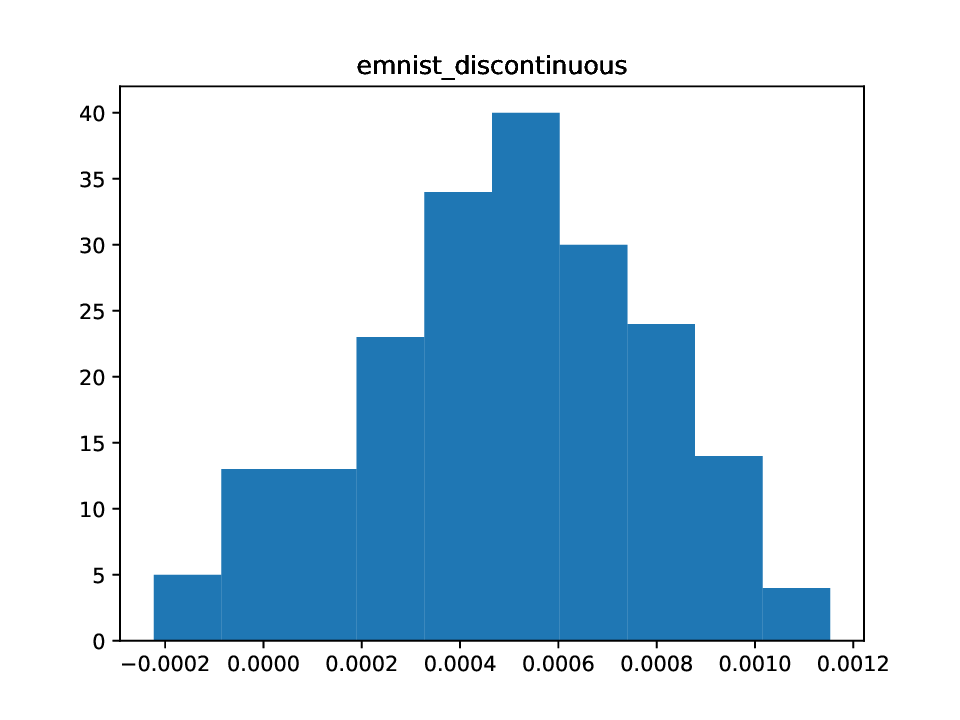}}
\caption{The distribution of the measured continuity bias for EMNIST.}
\label{fig:hist_emnist}
\end{figure}

\begin{figure}
    \centering
    \includegraphics{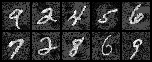}
    \caption{Adversarial examples of MNIST that fell inside the analytic polyhedra of the classifier that was trained on natural samples.}
    \label{fig:mnist_adv}
\end{figure}

\begin{figure}
    \centering
    \includegraphics{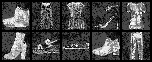}
    \caption{Adversarial examples of FMNIST that fell inside the analytic polyhedra of the classifier that was trained on natural samples.}
    \label{fig:fmnist_adv}
\end{figure}

\begin{figure}
\centering
\subfloat{\includegraphics[width=0.45\columnwidth]{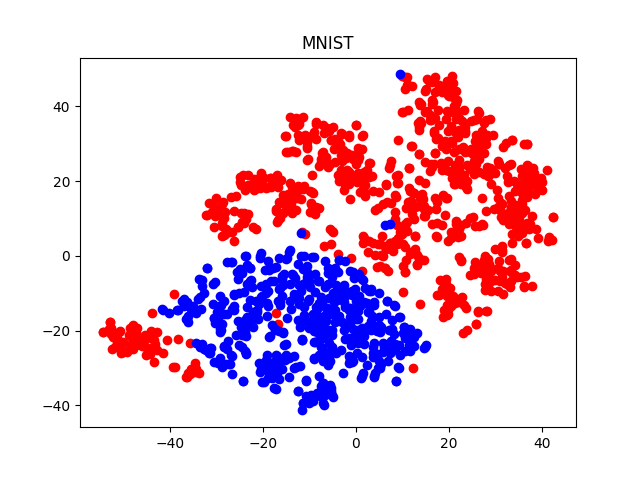}}
\subfloat{\includegraphics[width=0.45\columnwidth]{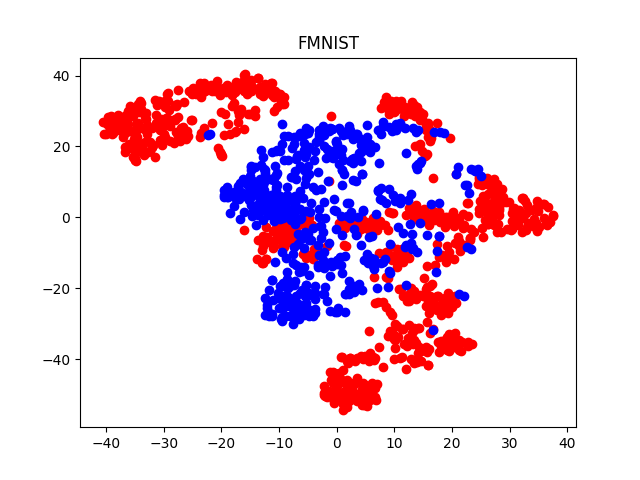}}
\caption{t-SNE embeddings of $S_\mathrm{ben}$ (blue) and $S_\mathrm{hard}$ (red) for MNIST and FMNIST.}
\label{fig:tsne}
\end{figure}

Finally, we present histograms of the measured continuity bias for the four datasets in Figures \ref{fig:hist_mnist}-\ref{fig:hist_emnist}. The vertical axis in the figures count the number of measurements that fell in a bin. Each measurement is computed for a batch of 32 samples from $S_\mathrm{hard}$ and 32 samples from $S_\mathrm{ben}$. For future reference, we present a sample of $S_\mathrm{hard}$ for MNIST and FMNIST in Figures \ref{fig:mnist_adv} and \ref{fig:fmnist_adv}. A depiction of separability of $S_\mathrm{hard}$ and $S_\mathrm{ben}$ using t-SNE is reported in Figure \ref{fig:tsne}.

\section{Details of implementations}
\label{sec:imp}
\begin{figure}
    \centering
    \resizebox{5cm}{!}{\begin{tikzpicture}

    % Define radii
    \def\rone{1.5}  % Inner circle
    \def\rtwo{3.25}  % Second circle (dashed)
    \def\rthree{5} % Third circle
    \def\rsquare{5.5} % Square size

    % Draw square
    \draw[thick] (-\rsquare, -\rsquare) rectangle (\rsquare, \rsquare);
    \node[below left] at (\rsquare, \rsquare) {\huge $\mathcal{X}$};

    % Fill the region between the innermost and outermost circles
    \fill[gray!20] (0,0) circle (\rthree);
    \fill[white] (0,0) circle (\rone); % Clear the innermost circle

    % Draw circles
    \draw[thick] (0,0) circle (\rone);  % Smallest circle
    \draw[dashed, thick] (0,0) circle (\rtwo);  % Dashed second circle
    \draw[thick] (0,0) circle (\rthree); % Third circle

    % Partitioning radial lines extending to the innermost circle
    \foreach \angle in {30, 150, 270} {
        \draw[thick] (\angle:\rthree) -- (\angle:\rone);
        \node[scale=1.2] at (\angle+5:{(2*\rthree+\rone)/3}) {$M$};
    }
    
    % Random small filled circles inside X_easy
    \foreach \x/\y in {1.5/0.5, -1.8/0.2, 0.7/1.2, -0.3/-1.1, 1.1/-1.5,-3.5/3.2, 3.8/-2.5, -4.2/-3.5, 2.5/4.1, -3.0/1.9} {
        \pgfmathsetmacro{\dist}{\x*\x + \y*\y}
        \ifthenelse{\lengthtest{\dist pt < 9pt}}
            {\fill[white,draw=black] (\x,\y) rectangle ++(7pt,7pt);}  % White inside Pi(h)
            {\fill[black] (\x,\y) rectangle ++(7pt,7pt);}  % Black outside Pi(h)
    }
    
    % Random small filled squares inside X square
    \foreach \x/\y in {-8.90e-1/-1.50e-0,-4.20e-1/-1.80e+0, 1.20e+0/1.60e+0,-1.80e-0/-3.70e-1, 1.90e+0/-7.00e-1, 7.00e-1/9.50e-1,-1.00e+0/1.70e+0, 1.70e+0/-6.60e-1,-1.10e+0/8.10e-1,-1.90e+0/1.30e+0, 8.10e-1/-1.50e-0} {
        \pgfmathsetmacro{\dist}{\x*\x + \y*\y}
        \ifthenelse{\lengthtest{\dist pt > 2.25pt}}
            {\fill[white,draw=black] (2*\x,2*\y) circle (4pt);}  % Black inside Pi(h)
            {\fill[black] (2*\x,2*\y) circle (4pt);}  % White outside Pi(h)
        
    }

    % Circle labels
    % \node[left] at (\rthree,0) {\huge $S_{easy}$};
    \node[above] at (0,\rtwo) {\huge $\partial\Pi(h)$};
    % \node[right] at (\rone,0) {\huge$S_{hard}$};

\end{tikzpicture}}
    \caption{A depiction of the proposed experiment for a classification problem with three categories. $\XX$ is the ambient domain and the gray area is the topologically non-trivial domain of benign samples. $M$ is the true accurate decision boundary that we want to robustly extend into $\XX$. The points that are circular are the benign samples and they are adversarial examples of the holomorphic hypothesis $h$ otherwise. The white circular points constitute $S_{\mathrm{easy}}$ and the white square points make up $S_{\mathrm{hard}}$. The boundary of the analytic polyhedra $\partial\Pi(h)$ of $h$ separates $S_{\mathrm{easy}}$ and $S_{\mathrm{hard}}$.}
    \label{fig:visual}
\end{figure}
In this section we will focus on giving a more practical perspective on previous sections and describe the details of the representation that we will be using to conduct our experiments. In our experiments, we are imagining a situation similar to Figure \ref{fig:visual}.

While the features proposed in theorem \ref{thm:varphi poly} are perfectly fine in lower dimensions, the dimensions of a typical learning problem is prohibitively large, and the count of the features would grow combinatorially with dimensions. A common solution to the issue of combinatorial explosion in machine learning is to assume independence between variables. These methods have been successfully applied to probabilistic graphical models in the past\cite[chapter~10]{bishop2006pattern}.

\begin{figure}
    \centering
    \includegraphics[width=0.5\columnwidth]{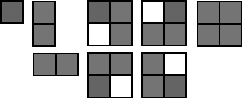}
    \caption{The templates that will be used in the experiments. These templates are chosen so that all overlapping templates of lower degree would take part in the model as well. The weight of the white pixels are 0 and otherwise they are 1.}
    \label{fig:temps}
\end{figure}

Here, we will adapt the strategy used by convolutional neural networks (CNN) to implement an image classifier. The idea is to use the Fourier basis and only compute features $\psi_\alpha$ for pixels that are close together. The overall architecture of the implementation is similar to a standard CNN with a few differences. First, it does not actually go deep. In other words, every filter is convolved with the image and then directly fed to the single dense layer at the end. Second, the weights of the filters are not learned during training and are fixed before training starts. The reason is that $\alpha\in\mathbb{N}^n$ and we would need to incorporate mixed-integer programming to do so. Consequently, we will call them templates instead of filters to emphasize this distinction. Each template will be reused with up to 5 pixels of dilation so that the model is able to capture correlations between far away pixels as well.

Given that we are trying to learn a hypothesis with minimal Dirichlet energy, we should incorporate all overlapping templates $\beta$ with $\|\beta\|_1<\|\alpha\|_1$ if we were to use $\alpha$ in the model. Otherwise, the model might attend to $\alpha$, when in fact the correlation was due to $\beta$. For example, if we were to include $e^{i(x_1+x_2)}$, then we must include $e^{ix_1}$ and $e^{ix_2}$ as well to make sure that the hypothesis with the smallest Dirichlet energy is learned. Finally, we will only consider templates with $\|\alpha\|_\infty\leq1$. We conjecture that features with $\|\alpha\|_\infty>1$ would not make any significant contribution to the model and will omit them. We have reported the templates that were used in the experiments in Figure \ref{fig:temps}.

To minimize the Dirichlet energy, we will \textit{tune} the eigenfunctions of the Laplace operator so that the implementation of weight decay in standard optimizers could be exploited to minimize the Dirichlet energy of the model.

\begin{definition}[tuning matrix]
Suppose that $\seq{\varphi_k}{1}{K}\subset\calDX$ is a set of features. The tuning matrix $\Sigma$ of $\seq{\varphi_k}{1}{K}$ is a $K\times K$ Hermitian matrix computed as:
\begin{equation}
\Sigma_{ij}=\volint{\calX}{\nabla\varphi_i(x)\cdot\nabla\varphi_j(x)}{x}.
\end{equation}
\end{definition}

In which $\calDX$ is the set of functions on $\calX$ with finite Dirichlet energy. Having computed the tuning matrix of $\seq{\varphi_k}{1}{K}$, it would be easy to compute the Dirichlet energy of $h$. The calculations would be more efficient if $\Sigma$ illustrates some structure.

\begin{definition}[weakly-harmonic features]
A set of features $\seq{\varphi_k}{1}{K}\subset \calDX$ is weakly-harmonic if its tuning matrix $\Sigma$ is the identity matrix $I$.
\end{definition}

\begin{theorem}
\label{thm:tuning}
Suppose that $\seq{\varphi_k}{1}{K}\subset\calDX$ is not weakly-harmonic. Then $\seq{\varphi_k}{1}{K}$ can be transformed into a set of weakly-harmonic features $\seq{\varphi^*_k}{1}{K}$ by:
\begin{equation}
\varphi^*(x)=\Sigma^{-\frac{1}{2}}\varphi(x).
\end{equation}
\end{theorem}

\begin{theorem}
\label{thm:weak harmonic energy}
Suppose that $\seq{\varphi_k}{1}{K}$ is weakly-harmonic and $h\in\calDX$ has a series representation in $\seq{\varphi_k}{1}{K}$. Then,
\begin{equation}
E[h]=\sum_{k=1}^{K}|w_k|^2.
\end{equation}
\end{theorem}

The final important part of the implementation is that we will attach a constant zero hypothesis to the classifier that other hypotheses have to beat before they get rewarded by the loss function during training. In other words, if there are 10 labels for the hypothesis to learn, we will add an extra label corresponding to $\Pi(h)$ and train the model as if there were 11 labels to learn. There is no need to add training samples for the extra label. The extra label is added to force the learning rule into finding a hypothesis that has pushed the natural samples outside its analytic polyhedra as much as possible.
%
%Exploiting the periodic nature of the Fourier bases, we will apply a different strategy to clipping for projecting the attacked sample inside $[0,1]^n$. Informally, when the attack hits the boundary of the domain, a clipping strategy would drag the sample along the boundary instead of letting the example leave $[0,1]^n$. In contrast, our strategy would be better described as the trajectory of a particle that bounces off of the boundary of the domain without losing any momentum. In other words, the path that the attack takes to get to the adversarial example is in reality a smooth path that leaves $[0,1]^n$. However, we can project the example back into $[0,1]^n$ by applying
%\begin{align}
%    \mathrm{proj}(x)=&
%    \begin{cases}
%        \mathrm{p}_1(x)&x>1\\
%        x&0\leq x\leq1\\
%        \mathrm{p}_0(x)&x<0
%    \end{cases},\\
%    \mathrm{p}_1(x)=&
%    \begin{cases}
%        \mathrm{fp}(x)&[x]\,\mathrm{even}\\
%        1 - \mathrm{fp}(x)&[x]\,\mathrm{odd}
%    \end{cases},\\
%    \mathrm{p}_0(x)=&
%    \begin{cases}
%        -\mathrm{fp}(x)&[x]\,\mathrm{even}\\
%        1 + \mathrm{fp}(x)&[x]\,\mathrm{odd}
%    \end{cases},\\
%    \mathrm{fp}(x)=&x-[x],
%\end{align}
%to each dimension of the generated adversarial example, and find a test sample inside $[0,1]^n$ for which $h$ would give the same prediction.

\section{Proofs of the Theorems}
% Use $\backslash${\tt{appendix}} if you have a single appendix:
% Do not use $\backslash${\tt{section}} anymore after $\backslash${\tt{appendix}}, only $\backslash${\tt{section*}}.
% If you have multiple appendixes use $\backslash${\tt{appendices}} then use $\backslash${\tt{section}} to start each appendix.
% You must declare a $\backslash${\tt{section}} before using any $\backslash${\tt{subsection}} or using $\backslash${\tt{label}} ($\backslash${\tt{appendices}} by itself
%  starts a section numbered zero.)}
\subsection*{Theorem \ref{thm:tuning}}
\begin{proof}
     \begin{align}
     \volint{\calX}{J_{\varphi^*}(x)J_{\varphi^*}(x)^T}{x}=&\Sigma^{-\frac{1}{2}}\Sigma\Sigma^{-\frac{1}{2}},\\
     =&I,
 \end{align}
 in which $J_{\varphi^*}(x)$ is the Jacobian matrix of $\seq{\varphi_k^*}{1}{K}$.
\end{proof}

\subsection*{Theorem \ref{thm:weak harmonic energy}}
\begin{proof}
 Given that $h$ has a series representation, we can compute its Dirichlet energy by:
 \begin{align}
     \volint{\calX}{\|\nabla h(x)\|^2}{x}=&\volint{\calX}{w^T J_\varphi(x)J_\varphi(x)^Tw}{x},\\
     =&w^T\volint{\calX}{J_\varphi(x)J_\varphi(x)^T}{x}w,
 \end{align}
 in which $J_\varphi(x)$ is the Jacobian matrix of $\seq{\varphi_k}{1}{K}$. But we know that
 \begin{equation}
     \volint{\calX}{J_\varphi(x)J_\varphi(x)^T}{x}=\Sigma=I.
 \end{equation}
 Thus, $E[h]=w^Tw$.
\end{proof}

\subsection*{Theorem \ref{thm:optimality}}
\begin{proof}
First, suppose that some real-valued $h:\Omega\to\R$ exists for which $\mathrm{sign}\big(h(z)\big)$ is PAC on $\Omega$ with respect to $M$ and define $f_0(z)=1,\Omega_0=\Omega\setminus M$. Then $\seq{(f_k,\Omega_k)}{k=1}{K}\cup\{(f_0,\Omega_0)\}$ is a holomorphic covering of $\Omega$ that is valid for a smoothing problem, and the second smoothing problem of $\seq{(h_k,T_{\omega_k})}{k=1}{K}$ is feasible by choosing
\begin{equation}
g'_k(z)=\frac{f_k(z)}{h(z)}.
\end{equation}
According to theorem \ref{thm:feasibility}, a holomorphic solution $\seq{g_k(z)}{k=1}{K}$ exists as well, and denote the smoothed holomorphic function by $f$. Since $g_k$ is nonvanishing, then $\mathbf{V}(f)=M$.

Now assume that some $f\in\mathcal{O}(T_\calX)$ exists for which $M=\mathbf{V}(f)$. Consequently, a continuous PAC classifier $h\in C(\calX)$ could be constructed,
\begin{equation}
    h(x)=|f(x)|t,
\end{equation}
in which $t:\Omega\to\{-1,+1\}$ is the discontinuous optimal classifier corresponding to $M$.
\end{proof}

\subsection*{Theorem \ref{thm:solution}}
\begin{proof}
We begin by recalling the Lagrangian of a weakly-harmonic learning problem in \eqref{eqn:weakly hamonic problem},
\begin{align}
    \calL&=\frac{1}{2}E[h]+\mathbb{E}_{X}[\loss(x;h,t)],\\
    &=\volint{\calX}{\frac{1}{2}\|\nabla h(x)\|^2+p(x)\loss(x,h)}{x},\\
    &=\volint{\calX}{L(x,h,\nabla h)}{x}.
\end{align}
According to variational calculus, the solution $h$ to the program satisfies the Euler-Lagrange equation,
\begin{equation}
    \frac{\delta L}{\delta h}(x)-\sum_{\|\alpha\|_1=1}\frac{\partial}{\partial x^\alpha}\frac{\delta L}{\delta h^{(\alpha)}}(x)=0,
\end{equation}
in which $\frac{\delta L}{\delta h^{(\alpha)}}$ is the variational derivative of $L$ with respect to $h^{(\alpha)}$. Substituting the Lagrangian inside the Euler-Lagrange equation we would have:
\begin{align}
    \frac{\delta L}{\delta h}(x)&=p(x)\loss_h(x,h),\\
    \frac{\delta L}{\delta h^{(\alpha)}}(x)&=h^{(\alpha)}\quad\|\alpha\|_1=1,\\
    0&=p(x)\loss_h(x,h)-\Delta h(x).
\end{align}
\end{proof}

\subsection*{Theorem \ref{thm:fundamental harmonic}}
\begin{proof}
We begin by first showing (2). Suppose that $\seq{\varphi_k}{1}{\infty}$ is the set of eigenfunctions of the Laplacian operator with eigenvalues $\seq{\lambda_k}{1}{\infty}$ and some function $u\in\calHX$. Then the Green function $G(x,x')$ satisfies the following PDE:
\begin{align}
    \Delta G(x,x')&=-\delta\big(\|x-x'\|\big),\\
    \frac{\partial G}{\partial\hat{n}}(x,x')&=0\quad x\in\partial\calX,
\end{align}
in which $\delta(\cdot)$ is the Dirac's delta. Since $\seq{\varphi_k}{1}{\infty}$ are the results of separation of variables, we can assume that $G(x,x')$ has a series representation:
\begin{equation}
    G(x,x')=\sum_{k=1}^\infty\frac{\varphi_k(x)\varphi_k(x')}{\lambda_k^2}.
\end{equation}
We can now use the divergence theorem to get
\begin{align}
    \volint{\calX}{\nabla\cdot \big(u\nabla G(x')\big)}{x}&=\surfint{\calX}{u\frac{\partial G(x')}{\partial\hat{n}}}{x},\\
    \volint{\calX}{\nabla\cdot \big(G(x')\nabla u\big)}{x}&=\surfint{\calX}{G(x')\frac{\partial u}{\partial\hat{n}}}{x},
\end{align}
in which $\mathrm{d}A(x)$ is the surface element of $\partial\calX$, and we have omitted the dependence of functions on $x$ for brevity,
and then apply the chain rule of divergence:
\begin{align}
    \nabla\cdot \big(u\nabla G(x')\big)&=\nabla u\cdot \nabla G(x')+u\Delta G(x'),\\
    \nabla\cdot \big(G(x')\nabla u\big)&=\nabla u\cdot \nabla G(x')+G(x')\Delta u,
\end{align}
and substitute the PDE definitions of $G(x')$ and $u$ to get
\begin{equation}
    u(x')=\surfint{\calX}{G(x')g}{x}-\volint{\calX}{G(x')f}{x}.
\end{equation} 
Thus, $u$ has a series representation in $\seq{\varphi_k}{1}{\infty}$.

Having proved (2), it is easy to show (1) as $\calHX$ can be exhausted by a countable increasing union of PAC learnable hypothesis spaces\cite[theorem~7.2]{shalev2014understanding}:
\begin{equation}
    \calHX=\bigcup_{K=1}^\infty\Big\{h\in H^2(\calX)\,|\,h=b+\sum_{k=1}^K w_k\varphi_k\Big\}.
\end{equation}

Finally, we can show (3) with the help of the PDE definition of $\varphi_k$, the divergence theorem and the chain rule of divergence to get
\begin{align}
    \volint{\calX}{\nabla \varphi_k\cdot \nabla \varphi_l}{x}&=\lambda_k^2\volint{\calX}{\varphi_k\varphi_l}{x},\\
    &=\lambda_l^2\volint{\calX}{\varphi_k\varphi_l}{x}.
\end{align}
Thus it must be that $\Sigma_{kl}=0$ when $\lambda_k\neq\lambda_l$. Moreover, in case $\lambda_k=\lambda_l$, $\Sigma_{kl}$ would again be zero when $\varphi_k$ and $\varphi_l$ differ in a separated eigenfunction because eigenfunctions of the resulting Sturm-Liouville ordinary differential equations are orthogonal.
\end{proof}

\subsection*{Theorem \ref{thm:varphi poly}}
\begin{proof}
We begin by assuming that any eigenfunction can be separated in $x$,
\begin{equation}
    \varphi_\alpha(x)=\prod_{j=1}^n\Phi(x_j;\alpha_j).
\end{equation}
Thus,
\begin{align}
    \Delta \varphi_\alpha(x)&=-\lambda_\alpha^2\varphi_\alpha(x),\\
    \varphi_\alpha(x)\sum_{j=1}^n\frac{\Phi^{(2)}(x_j;\alpha_j)}{\Phi(x_j;\alpha_j)}&=-\lambda_\alpha^2\varphi_\alpha(x),\\
    \sum_{j=1}^n\frac{\Phi^{(2)}(x_j;\alpha_j)}{\Phi(x_j;\alpha_j)}&=-\lambda_\alpha^2.
\end{align}
Now we can separate the variables and reach the Sturm-Liouville ordinary differential equations
\begin{align}
    \Phi^{(2)}(x_j;\alpha_j)+\alpha_j^2\Phi(x_j;\alpha_j)&=0, \quad \|\alpha\|_2^2=\lambda_\alpha^2,\\
    \Phi^{(1)}(0;\alpha_j)=\Phi^{(1)}(\pi;\alpha_j)&=0.
\end{align}
Then,
\begin{equation}
    \Phi(x_j;\alpha_j)=c_1\cos(\alpha_j x_j)+c_2\sin(\alpha_j x_j).
\end{equation}
However, given the boundary conditions we can see that $c_2=0$.

The corresponding diagonal element of $\Sigma$ is:
\begin{align}
    \Sigma_{jj}&=\|\alpha_j\|_2^2\volint{\omega}{|\varphi_j(x)|^2}{x},\\
    &=\|\alpha_j\|_2^2\volint{\omega}{\prod_{l=1}^n\cos^2(\alpha_{jl}x_l)}{x},\\
    &=\frac{\pi^n}{2^n}\|\alpha_j\|_2^2.
\end{align}
After scaling $\Sigma$ with $\frac{2^n}{\pi^n}$, the proof is complete.
\end{proof}

%\subsection*{Lemma \ref{lem:varphi neural}}
%\begin{proof}
%Consider the product to sum formula of cosines,
%\begin{equation}
%    \cos(x_j)\cos(x_k)=\frac{1}{2}\big(\cos(x_j+x_k)+\cos(x_j-x_k)\big).
%\end{equation}
%Thus, we can transform $\varphi_\alpha$ by repeated application of the product to some formula to a summation of $2^{n-1}$ cosine functions,
%\begin{multline}
%    \prod_{j=1}^n\cos(\alpha_j x_j)=\frac{1}{2^{n-1}}\big(\cos(\alpha_1x_1+\cdots+\alpha_nx_n)+\cdots\\
%    +\cos(\alpha_1x_1-\cdots-\alpha_nx_n)\big).
%\end{multline}
%To make the expression invariant to the choice of the starting index, we exploit the evenness of the cosine function and add $2^{n-1}$ more equal cosine functions with negated inputs to the expression to get to the theorem. We should emphasize that there is no need to compute all $2^n$ terms because $\cos(\alpha_jx_j)=1$ when $\alpha_j=0$ and $\alpha$ is sparse in practice.
%\end{proof}

\subsection*{Theorem \ref{thm:tw}}
\begin{proof}
    We will continue from where we left off in the proof of theorem \ref{thm:varphi poly}. Now that we have an extra dimension for every real dimension, we only have to separate each $\Phi$ again between $x_j$ and $y_j$,
    \begin{equation}
        \Psi(x_j,y_j;\alpha_j)=\Phi(x_j;\alpha_j)\Theta(y_j;\alpha_j),
    \end{equation}
    and then solve for $\Delta\Psi(x_j,y_j;\alpha_j)=0$. Consequently,
    \begin{equation}
        \Theta(y_j;\alpha_j)=c_1\cosh(\alpha_jy_j)+c_2\sinh(\alpha_jy_j).
    \end{equation}
    Assuming $\frac{\partial\Theta}{\partial y_j}(0)=0$, we can see that $c_2=0$. However, another function $\Psi^\dagger$ exists for which $\Psi^\dagger(x_j,0;\alpha_j)=0$,
    \begin{equation}
        \Psi^\dagger(x_j,y_j;\alpha_j)=\sin(\alpha_jx_j)\sinh(\alpha_jy_j),
    \end{equation}
    called the conjugate harmonic of $\Psi(x_j,y_j;\alpha_j)$ and we have
    \begin{equation}
        \cos(\alpha_jz_j)=\Psi(x_j,y_j;\alpha_j)+i\Psi^\dagger(x_j,y_j;\alpha_j),
    \end{equation}
    which satisfies the Cauchy-Riemann equations and thus is holomorphic.
\end{proof}

%\subsection*{Theorem \ref{thm:psi}}
%\begin{proof}
%    The first part of the theorem is a direct consequence of the existence of a biholomorphic map between $U_\omega$ and $D^n(0,1)$ and the fact that $z^\alpha$ is an orthogonal set of basis functions for the intersection of $\mathcal{O}(D^n(0,1))$ and $L^2(D^n(0,1))$, which is a Hilbert space called the Bergman space $A^2(D^n(0,1))$ of the poly-disk\cite[p.~60]{krantz2001function}. The inner product of $A^2$ uses the usual inner product of $\setC^n$ which conjugates the left side of the inner product:
%    \begin{equation}
%        z_1\cdot z_2=z_1^Hz_2.
%    \end{equation}
%    
%    We can compute $\Sigma_{kl}$ as:
%    \begin{align}
%        \Sigma_{kl}&=\volint{\omega}{\nabla \psi_{\alpha_k}(x)\cdot\nabla \psi_{\alpha_l}(x)}{x},\\
%        &=\alpha_k\cdot\alpha_l\volint{\omega}{e^{i(\alpha_l-\alpha_k)\cdot x}}{x},\\
%        &=
%        \begin{cases}
%            (2\pi)^n\|\alpha_k\|_2^2&\quad k=l\\
%            0&\quad \mathrm{o.w.}
%        \end{cases}.
%    \end{align}
%    We then scale $\Sigma$ with $(2\pi)^{-n}$ and the theorem follows.
%\end{proof}

\subsection*{Theorem \ref{thm:feasibility}}
\begin{proof}
    We will proceed by turning the smoothing problems into Cousin problems\cite[chapter~6]{krantz2001function}.
    
    Consider the first smoothing problem, and define functions $g_{lk}:\Omega_k\cap\Omega_l\to\setC$ such that
    \begin{equation}
        g_{kl}(z)=g_l(z)-g_k(z) \quad z\in \Omega_k\cap \Omega_l.
    \end{equation}
    It can be checked that
    \begin{align}
        g_{kl}(z)+g_{lk}(z)&=0 \quad z\in \Omega_k\cap \Omega_l,\\
        g_{jk}(z)+g_{kl}(z)+g_{lj}(z)&=0\quad z\in \Omega_j\cap\Omega_k\cap\Omega_l.
    \end{align}
    Thus $\seq{g_{kl}}{k,l=1}{K}$ is valid for a first Cousin problem\cite[page~247]{krantz2001function}. The first Cousin problem is always solvable by continuous functions\cite[proposition~6.1.7]{krantz2001function}. Furthermore, if $\Omega$ is a domain of holomorphy, the first Cousin problem is also solvable by holomorphic functions\cite[proposition~6.1.8]{krantz2001function}. This will conclude the proof of the first part of the theorem.

    Now consider the second smoothing problem, and define functions $g_{lk}:\Omega_k\cap\Omega_l\to\setC$ such that
    \begin{equation}
        g_{kl}(z)=\frac{g_l(z)}{g_k(z)} \quad z\in \Omega_k\cap \Omega_l.
    \end{equation}
    It can be checked that
    \begin{align}
        g_{kl}(z)g_{lk}(z)&=1 \quad z\in \Omega_k\cap \Omega_l,\\
        g_{jk}(z)g_{kl}(z)g_{lj}(z)&=1\quad z\in \Omega_j\cap\Omega_k\cap\Omega_l.
    \end{align}
    Thus $\seq{g_{kl}}{k,l=1}{K}$ is valid for a second Cousin problem\cite[page~247]{krantz2001function}. The second Cousin problem can be solved by holomorphic functions on $\Omega$ if and only if it can be solved just continuously\cite[proposition~6.1.11]{krantz2001function}.
\end{proof}
 

% \end{IEEEbiography}



% \vfill

\end{document}